\documentclass{article}

\PassOptionsToPackage{square, numbers}{natbib}

\usepackage[preprint]{neurips_2023}

\usepackage[utf8]{inputenc} % allow utf-8 input
\usepackage[T1]{fontenc}    % use 8-bit T1 fonts
\usepackage{hyperref}       % hyperlinks
\usepackage{url}            % simple URL typesetting
\usepackage{booktabs}       % professional-quality tables
\usepackage{amsfonts}       % blackboard math symbols
\usepackage{nicefrac}       % compact symbols for 1/2, etc.
\usepackage{microtype}      % microtypography
\usepackage{xcolor}         % colors
\usepackage{graphicx}
\usepackage{amsmath}
\usepackage{subcaption}

\title{Learning and Generalizing Polynomials in Simulation Metamodeling}

\author{%
  Jesper Hauch, Christoffer Riis, Francisco C. Pereira \\
  DTU Management \\
  Technical University of Denmark \\
  2800 Kgs. Lyngby, Denmark \\
}

\DeclareUnicodeCharacter{2212}{-}
\begin{document}

\bibliographystyle{abbrvnat}
\maketitle

\begin{abstract}
The ability to learn polynomials and generalize out-of-distribution is essential for simulation metamodels in many disciplines of engineering, where the time step updates are described by polynomials. While feed forward neural networks can fit any function, they cannot generalize out-of-distribution for higher-order polynomials. Therefore, this paper collects and proposes multiplicative neural network (MNN) architectures that are used as recursive building blocks for approximating higher-order polynomials. Our experiments show that MNNs are better than baseline models at generalizing, and their performance in validation is true to their performance in out-of-distribution tests. In addition to MNN architectures, a simulation metamodeling approach is proposed for simulations with polynomial time step updates. For these simulations, simulating a time interval can be performed in fewer steps by increasing the step size, which entails approximating higher-order polynomials. While our approach is compatible with any simulation with polynomial time step updates, a demonstration is shown for an epidemiology simulation model, which also shows the inductive bias in MNNs for learning and generalizing higher-order polynomials.
\end{abstract}

\section{Introduction}
% Inductive biases
% Simulations and simulation metamodeling
% Polynomials
Inductive learning is a fundamental task of machine learning models describing the process where patterns are induced from data. In the field of machine learning, multilayer feed forward neural networks (FFNNs) have been shown to be powerful inductive learners, as they are universal approximators capable of approximating any measurable function to any desired degree of accuracy \cite{HORNIK1989359}. However, the high flexibility of neural networks makes them prone to overfitting the data on which they are trained, indicating that they cannot properly generalize to unseen data. There exist approaches to avoid overfitting during training \cite{Overfitting}, but even when these are successfully applied, models can still struggle to generalize to unseen data, especially when the unseen data come from a different distribution than the training data. In this case, additional biases are critical for neural networks to generalize, since an unbiased program, which uses training data as its only source of information, cannot outperform programs that use rote learning \cite{Mitchell80}. To enable neural networks to generalize to unseen out-of-distribution data, inductive biases can be introduced by making reasonable assumptions about the underlying patterns of the data on which induction is performed.

% Bridge to simulations and metamodeling
Inductive learning is closely related to how progress is made in the field of natural sciences, where empirical evidence from experimentation and observation is used to understand the world. More specifically, the aim is to understand a process and its underlying mechanisms to model the outcome in different situations, for example, using simulation. Simulation models are a popular tool for exploring the behavior of a process over time, given a set of parameters and state variables. The state variables describe the state of the process at some given time step, and their behavior over time are expressed as differential equations in many disciplines of engineering. As a result, a simulation run is performed by solving the system of differential equations that describes the process \cite{SimBook}. Unfortunately, solving this system for high-complexity processes can be computationally expensive to an extent that renders simulation almost useless in practice. An approach to this problem is to use metamodeling, where a simpler mathematical model is introduced to approximate the behavior induced by the differential equations of the computationally expensive simulation model. In discrete event simulation, these differential equations can be expressed as polynomials that relate the current values of state variables to the past values. The order of these polynomials is determined by the number of discrete time steps that are approximated at a time, where approximating multiple time steps entails approximating higher-order polynomials. Therefore, a simulation metamodel should be able to approximate both low- and high-order polynomials to be successful in many engineering applications. Neural networks are popular metamodels for this purpose, as FFNNs should be able to approximate any function if enough neurons and layers are used \cite{HORNIK1989359}.

\begin{figure}
\centering
    \hspace{0.05\textwidth}
    \begin{subfigure}[t]{0.4\textwidth}
    \centering
    \includegraphics[width=\textwidth]{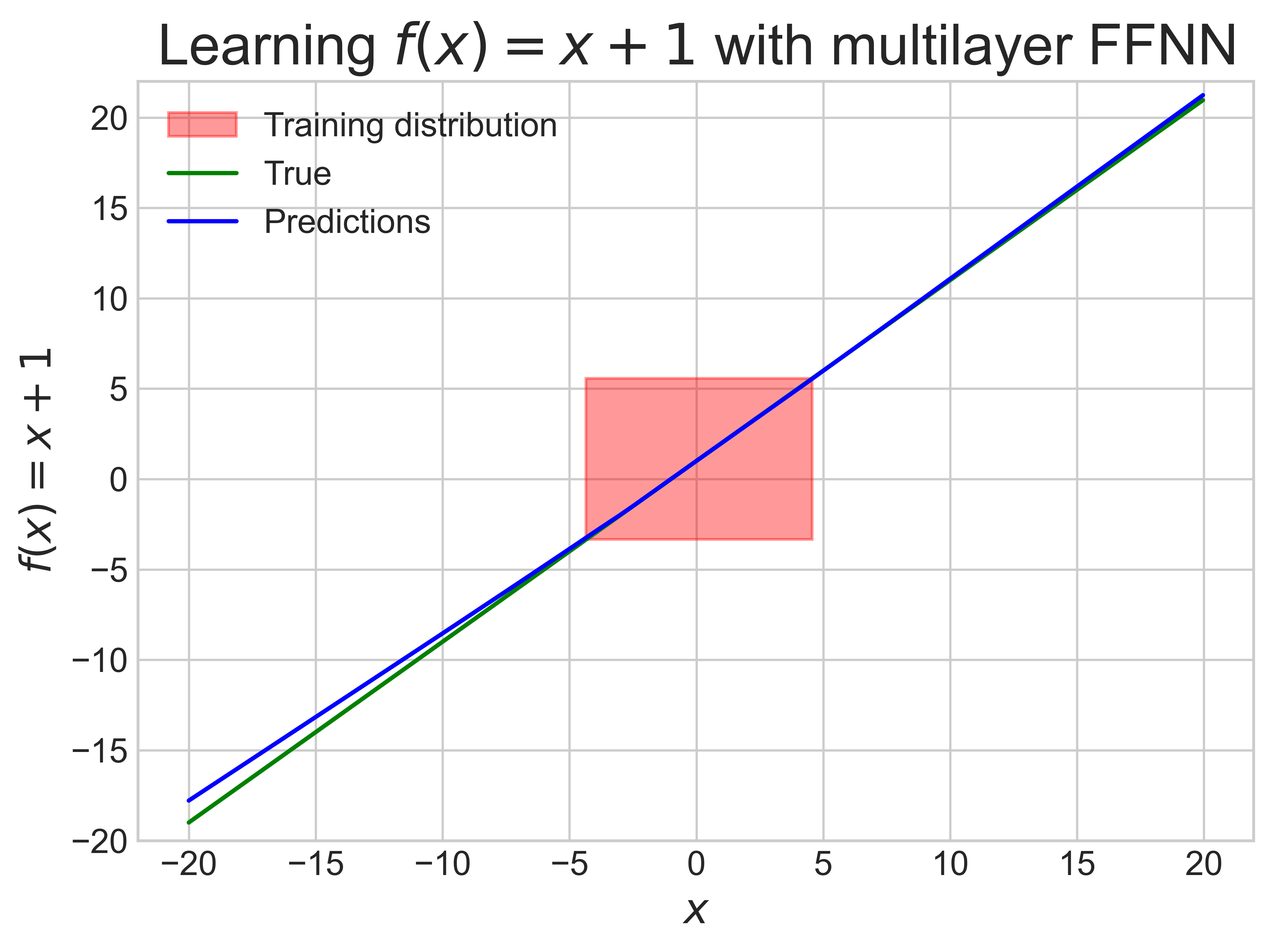}
    \caption{In and out-of-distribution performance for $f(x)=x+1$.}
    \label{subfig:intro-x}
    \end{subfigure}
    \hfill
    \begin{subfigure}[t]{0.4\textwidth}
    \centering
    \includegraphics[width=\textwidth]{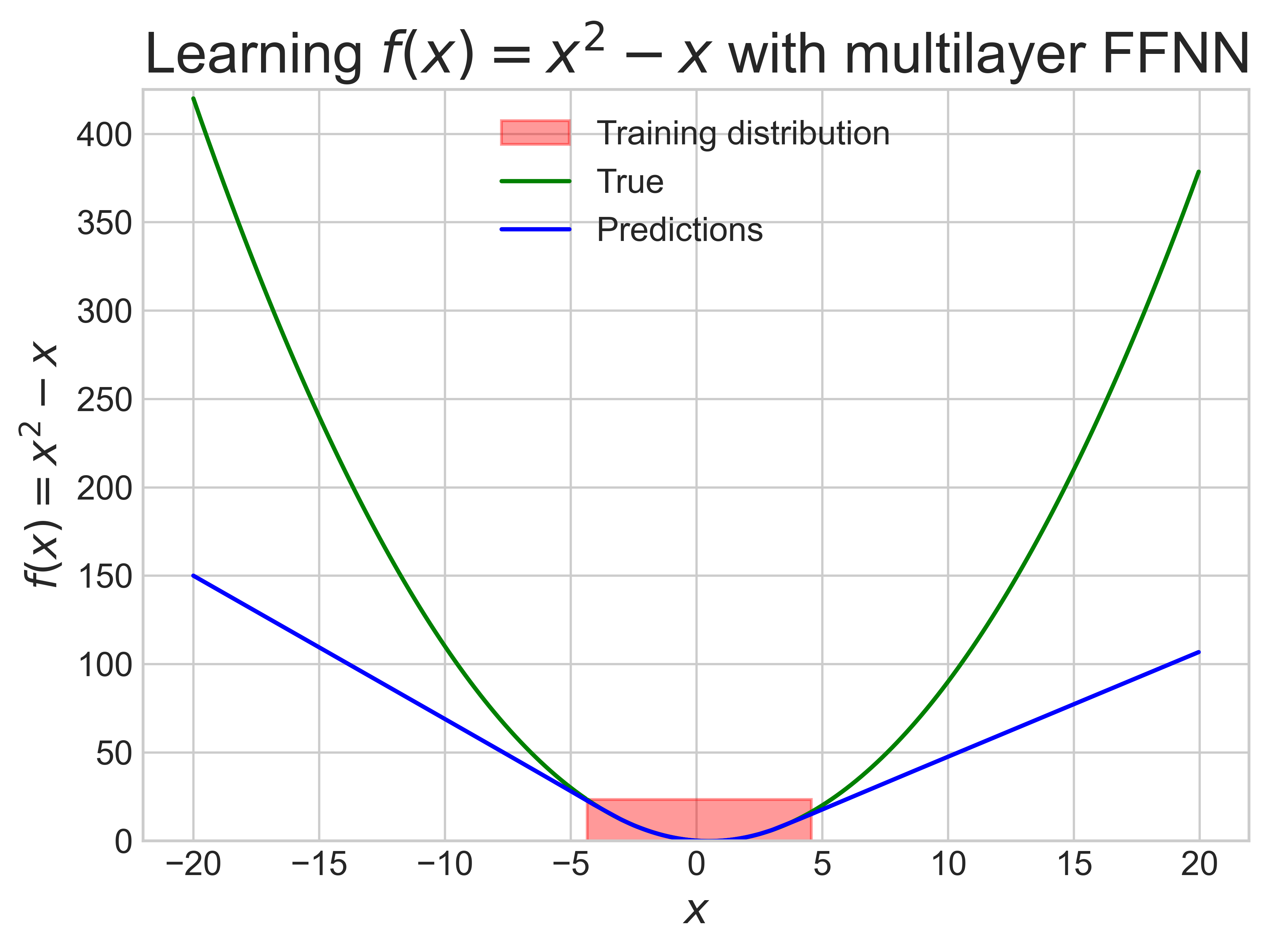}
    \caption{In and out-of-distribution performance for $f(x)=x^2-x$.}
    \label{subfig:intro-x2}
    \end{subfigure}
    \hspace{0.05\textwidth}
\caption{Multilayer FFNN with ReLU activation learning low-order polynomials with samples from $\mathcal{N}(0,1)$ distribution.}
\label{fig:intro_motivation}
\end{figure}

%In science, many processes are described and modeled using polynomial functions, since they can approximate any continuous function defined on a closed interval to any desired degree of closeness \cite{Weierstrass-Theorem}.
% Bridge to polynomials
As an example, consider the two simple low-order polynomial functions $f(x)=x+1$ and $f(x)=x^2-x$. These polynomials are approximated using a multilayer FFNN with ReLU activation that learns from samples drawn from a conditional standard normal distribution. If the multilayer FFNN has learned the underlying polynomials, it should be able to generalize to samples outside of the training distribution. The polynomial $f(x)=x+1$ is closely approximated within the training distribution, as seen in Figure \ref{subfig:intro-x} by the near perfect fit between the lines in the red area. As samples further away from the training distribution are used, the error between the true values and predictions starts to grow but remains relatively low. Likewise, for the quadratic polynomial $f(x)=x^2-x$ in Figure \ref{subfig:intro-x2}, the fit is also near perfect within the training distribution, but errors begin to occur as soon as samples outside the training distribution are used. When samples are used that lie even further away from the training distribution, the error between the true values and predictions is large.

% Contributions
The example shows that FFNNs, which are architectures without inductive biases, can learn polynomials in-distribution but struggle with out-of-distribution generalization, especially for polynomials of second-order. It is reasonable to assume that the polynomials describing real-world processes are much more complex than the simple ones in the example, which implies that FFNNs perform even worse when modeling these processes out-of-distribution. Therefore, in this thesis, we focus on the development of neural network simulation metamodels that can learn polynomials and generalize out-of-distribution. The following contributions are made:
\begin{itemize}
    \item Collect and repurpose multiplicative neural network (MNN) architectures, which are building blocks that can be used recursively to model polynomials of arbitrary order. This thesis proposes the PDCLow architecture based on MNNs in the literature.
    \item Performance evaluation is done out-of-distribution from the training distribution to test whether MNNs are able to generalize. Across all experiments, MNNs outperform baseline models in out-of-distribution tests, and their performance in validation is true to their performance in out-of-distribution tests.
    \item A novel metamodeling approach for discrete event simulations with time step updates described by polynomials. The approach is demonstrated in this work for an epidemiology simulation model.
    \item The code is available for open source use at \url{https://github.com/jesperhauch/polynomial_deep_learning}.
\end{itemize}

The remaining sections of the paper are structured as follows. In Section \ref{sec:literature-review}, the relevant literature is reviewed. Section \ref{sec:methodology} describes how higher-order outputs can be produced for polynomial learning and presents the proposed simulation metamodeling approach. The experimental setup and the experiments themselves are thoroughly described in Section \ref{sec:experiments}, while their results are evaluated in Section \ref{sec:results} with an emphasis on out-of-distribution generalization. Limitations and ideas for future work are discussed in Section \ref{sec:discussion} before the paper is concluded in Section \ref{sec:conclusion}.

\section{Literature Review}\label{sec:literature-review}
This section reviews the literature on simulation metamodeling and the literature that bridges the gap between neural networks and polynomials. The first section describes the machine learning models used in the metamodeling literature, where emphasis is placed on their generalization capabilities. Some of this literature is implicitly concerned with polynomial fitting, which creates a natural transition to the second section that focuses solely on approaches that explicitly approximate polynomials. The final section explains the development and use of polynomial neural networks.

\subsection{Machine learning metamodels}
% Kriging
% Random Forests
% Bayesian Networks
% Neural Networks
This section reviews the use of machine learning models that approximate the input-to-output function of simulators, also known as \emph{simulation metamodels}. Focus is placed on the advantages and disadvantages of using certain machine learning models for metamodeling, especially in terms of their out-of-distribution generalization capabilities.

% Random Forest: Ensemble
Random Forest (RF) is a machine learning model that generates a set of decision trees to make a prediction collectively. They are capable of fitting non-linear data using a small number of samples, which can be useful when only a limited amount of data from the simulator is available \cite{RFSurrogate}. Diverse applications have seen RF act as a decision support in design optimization \cite{RFSurrogate, RFGenetic} and as predictors of soil emissions \cite{RFSoil} and drainage fraction in water resource management \cite{RFDrainage}. RF models are powerful but can be computationally expensive to train, depending on the size of the ensemble and the depth of the decision trees. Furthermore, RF models can be undesirable for metamodeling, as they do not generalize well beyond training data if testing is performed out-of-distribution \cite{RFDrainage}. 

% bayesian networks
The lack of generalization capabilities is also a problem for Bayesian networks, despite the widespread assumption that they are good at out-of-distribution detection \cite{OODBN}. Nevertheless, Bayesian networks have been applied in metamodeling with great success in terms of interpretability and computational efficiency. In discrete event simulation, Bayesian networks can track the probability distribution of the simulation state over the duration of the simulation. This enables in-depth analysis in individual time steps and what-if analyzes, where state variables can be fixed in a certain time step to investigate their influence \cite{POROPUDAS2011644}. Although Bayesian networks are highly interpretable models, implementing them successfully can be tedious. Their structure is determined by the cause-effect relationship between variables, which is difficult to assess for complex processes \cite{UUSITALO2007312}, and cyclical relationships between variables cannot be dealt with \cite{JensenNielsen}. These factors, among others, result in Bayesian networks not being commonly used for metamodeling.

% Gaussian Process
A common machine learning metamodel is the Gaussian process, which is also referred to as a kriging model in the literature. Ordinary kriging can be applied in deterministic simulation \cite{Kriging}, but extensions have enabled applications in stochastic simulation \cite{StochasticKriging}. Unfortunately, kriging models scale poorly, since an unknown correlation function has to be approximated with maximum likelihood estimation \cite{Kriging}. As a result, the number of input variables and the amount of training data must be low to reduce the computation time. Although the latter is an acceptable limitation in metamodeling (due to often slow simulation tools), the number of input variables may need to be high depending on the process at hand, making this a challenge. On the other hand, a benefit of kriging models compared with other approaches is that the uncertainty of the predictor is quantified by default, which facilitates transparency of performance if kriging is applied in out-of-distribution scenarios in practice. As an extension to the default uncertainty estimation, the work of \citet{SKPIs} even combines stochastic kriging with neural networks to provide prediction intervals that quantify the uncertainty of a prediction with greater precision.

% Neural networks
Neural networks are also widely used for metamodeling because of their flexible architectures. The work by \citet{Kasim_2021} proposes a method based on neural architecture search, which enables the acceleration of scientific simulations up to 2 billion times using limited training data \cite{Kasim_2021}. Simpler neural networks, such as FFNNs, have also been successfully applied to accelerate simulation time \cite{SHAHRIARI2020235, RobustSim}, but their out-of-distribution generalization capabilities are rarely explored. However, the use of convolutional neural networks has shown promising results in generalization to unseen data \cite{CNNMeta}. This is likely to be attributable to the inductive biases of convolutional neural networks, which are useful for computer vision applications.

\subsection{Polynomial Approximation}
% Polynomial fitting
%% NN-poly
%% Mapping polynomial fitting into feedforward neural networks
The machine learning metamodels covered in the previous section are implicitly learning the often underlying polynomial input-to-output functions of simulators. In contrast, this section reviews the literature explicitly learning polynomial functions. Recent work by \citet{NN-poly} proposes how common neural networks can adhere to physical laws by representing them as Taylor polynomials of arbitrary order and placing constraints on their state predictions. While their work assumes that the true polynomials that model dynamical systems are unknown from a learning perspective, the work of \citet{Map-poly-fitting} attempts to approximate the terms of known polynomials with neural networks. By assuming that the polynomial is known, they design neural network architectures that are specific for a single polynomial term, e.g. approximating the third polynomial term alone. Unfortunately, their architectures are isolated to one-variable and two-variable cases up to the third-order. However, designing neural networks to approximate \emph{known} mathematical functions is the first step toward understanding how neural networks can be designed to solve \emph{unknown} problems \cite{Map-poly-fitting}.

\subsection{Polynomial Neural Networks}\label{sec:lr-pnn}
\begin{figure}[ht]
    \centering
    \includegraphics[width=0.4\textwidth]{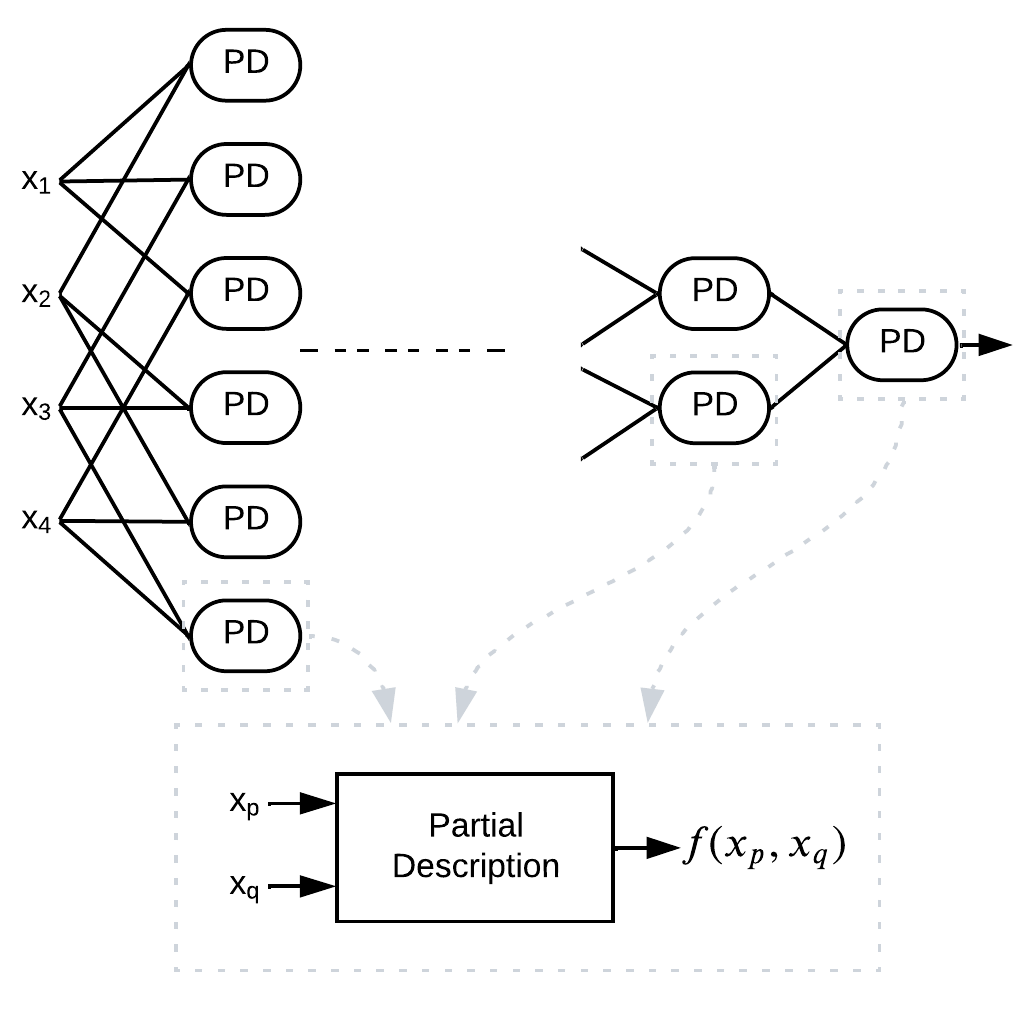}
    \caption{Basic polynomial neural network (PNN) structure \cite{poly-oh}. The partial descriptions (PDs) are subsets of the input variables.}
    \label{fig:pnn}
\end{figure}

% Polynomial Activations
%% Polynomial neural networks architecture: analysis and design
The combination of polynomials and neural networks happened in the late 1960s with the Group Method of Data Handling algorithm, which can automatically determine the best performing neural network structure for a specific application. The structure is generated by selecting a subset of two input variables in each neuron, termed partial descriptions (PDs), and treating them as quadratic regression polynomials \cite{Ivakhenko-GMDH}. The highly systematic design approach means that the layers and neuron connections change dynamically during training. Unfortunately, self-organizing networks following this approach tend to generate complex polynomials for simple systems, produce overly complex models, and lack flexibility if there are fewer than three input variables \cite{poly-oh}. These problems are alleviated in another approach to self-organizing networks called polynomial neural networks (PNNs) \cite{poly-oh}, where the PDs can have a different number of input variables and exploit polynomials of different order. The structure of the PNN is selected based on the number of input variables and the order of the PDs in every layer. In the basic PNN structure shown in Figure \ref{fig:pnn}, the number of input variables is the same in every layer, while the modified PNN structure can have varying numbers of input variables from layer to layer. These two structures are applied in simulations of gas furnaces, sewage treatment, and non-linear static systems, where they outperform baseline models in mean squared error and stability index. The stability index expresses the performance deterioration of the model in the test data and is a good indicator of the approximation and generalization capabilities of the PNN. As PNNs are self-organizing, their structure is automatically determined for an application, whereas another study considers PNNs as FFNNs equipped with a polynomial activation function or transfer function to accommodate polynomial approximation \cite{kileel2019expressive}.

\section{Methodology}\label{sec:methodology}
This section concerns the methodology that comprises the main contributions of this thesis. The first section describes the multiplicative neural networks (MNNs) used to produce higher-order outputs from input. The second section covers the simulation metamodeling approach, compatible with simulations where state updates are described by polynomials.

\subsection{Producing Higher-Order Outputs}\label{subsec:producing-higher-outputs}
The introduction of this paper highlights why unbiased estimators, such as an FFNN, require inductive biases to generalize beyond the training data. Our introductory example in Figure \ref{fig:intro_motivation} shows how a multilayer FFNN is able to generalize well for a first-order polynomial, but struggles for a quadratic function. In contrast to the unbiased architecture of such models, this section concerns architectures aiming at producing higher-order outputs. These architectures are termed multiplicative neural networks (MNNs) because of their use of multiplications to raise the input to higher-order output. 

\begin{figure}[ht]
\centering
\includegraphics[width=0.4\textwidth]{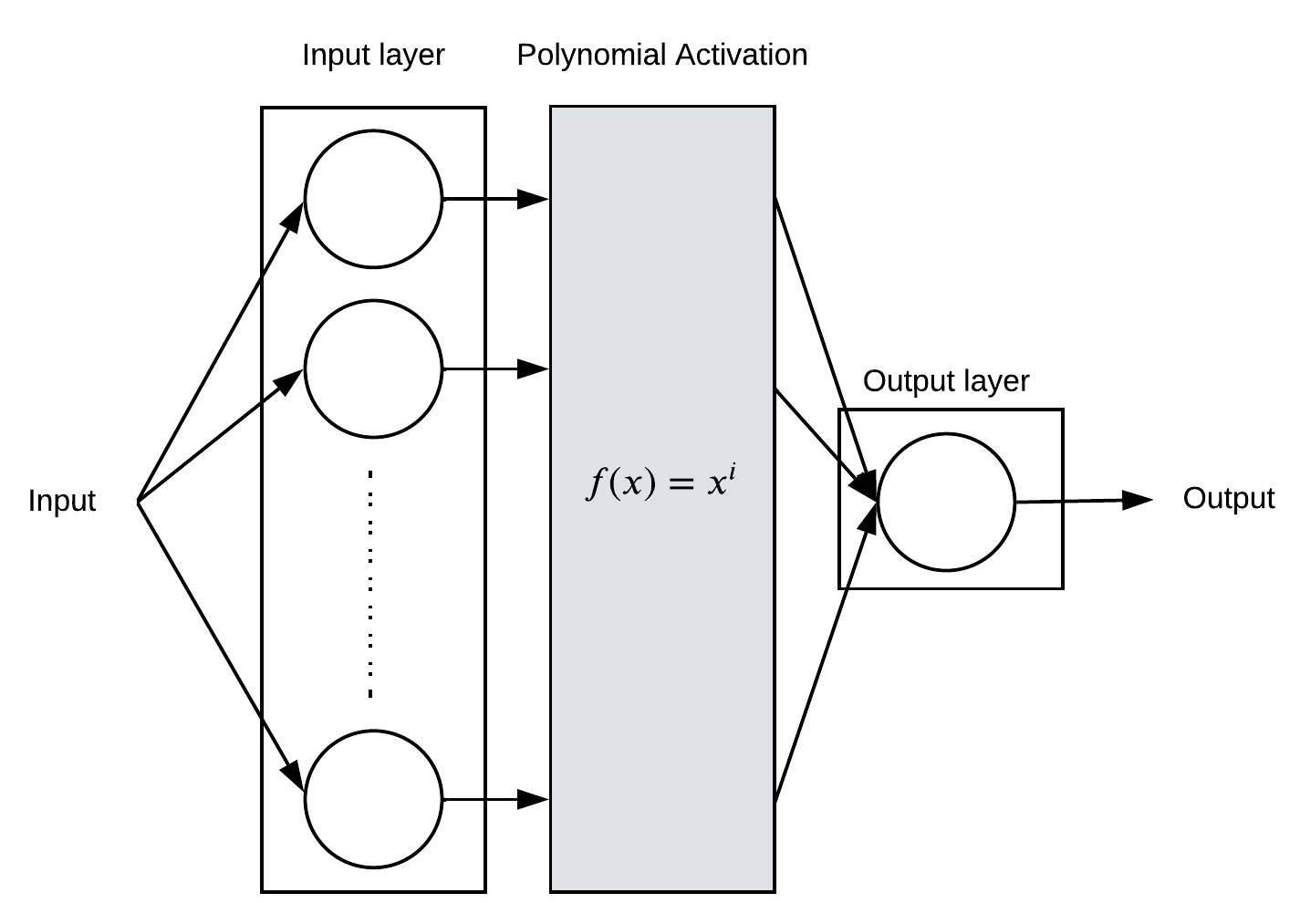}
\caption{Polynomial Activation Neural Network (PANN). PANN is a FFNN with polynomial activation, which raises input to the \textit{i}\textsuperscript{th} order.}
\label{fig:pann}
\end{figure}

The first architecture is highly inspired by the work on deep PNNs \cite{kileel2019expressive}, which defines PNNs as FFNNs with a polynomial activation function. The polynomial activation function simply applies a selected exponent to the input, as shown in Figure \ref{fig:pann}, to facilitate the approximation of higher-order polynomials. However, this architecture only facilitates fitting the highest-order term in a polynomial and neglects lower-order terms, which is problematic depending on the polynomial at hand.

\begin{figure}[ht]
\centering
\includegraphics[width=0.6\textwidth]{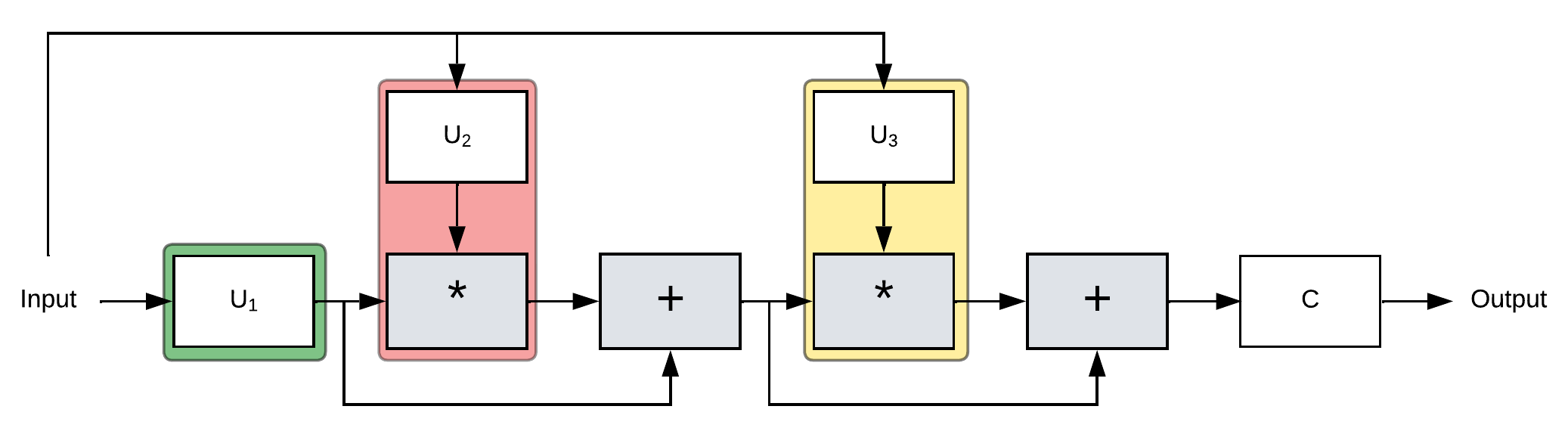}
\caption{Third-order Coupled CP (CCP) model \cite{pi_nets}. The architecture can be used recursively to achieve higher-order output than shown here. The output of the green area is of first-order, red area is second-order, and yellow area is third-order.}
\label{fig:ccp}
\end{figure}

While the PANN uses a polynomial activation function to produce higher-order outputs, the Coupled CP decomposition (CCP) from the $\Pi$-Nets paper \cite{pi_nets} uses a special kind of skip connection instead. Their work focuses on applications that require convolutional layers in the CCP architecture, such as image generation and classification, whereas this thesis models input-output functions using linear layers. A visualization of a CCP architecture that can generate third-order output is found in Figure \ref{fig:ccp}, where the linear layers are shown as white boxes, the mathematical operations are shown as gray boxes, and the different colored areas indicate the polynomial order of the output produced by that area. The green area produces first-order output using the linear layer $U_1$, which performs a linear transformation of the input. The output of the green area is multiplied by the first-order output of $U_2$ in the red area, which raises the output to the second-order. To preserve a term of first-order, the output from the green area is added to the second-order output via a skip connection. The concept of using the same first-order output to both represent the first-order term and raise the input to the second-order is called weight sharing, since the weights in $U_1$ are shared between these two use cases. When the weights are shared, the number of parameters that need to be trained is reduced, increasing the computational efficiency of the training process. The aforementioned procedure for the red area is repeated for the yellow area with the output of the red area, where a third-order output is generated, and is repeatable to generate output of any desired order.

\begin{figure}[ht]
\centering
\hspace{0.05\textwidth}
\begin{subfigure}[t]{0.4\textwidth}
    \centering
    \includegraphics[width=\textwidth]{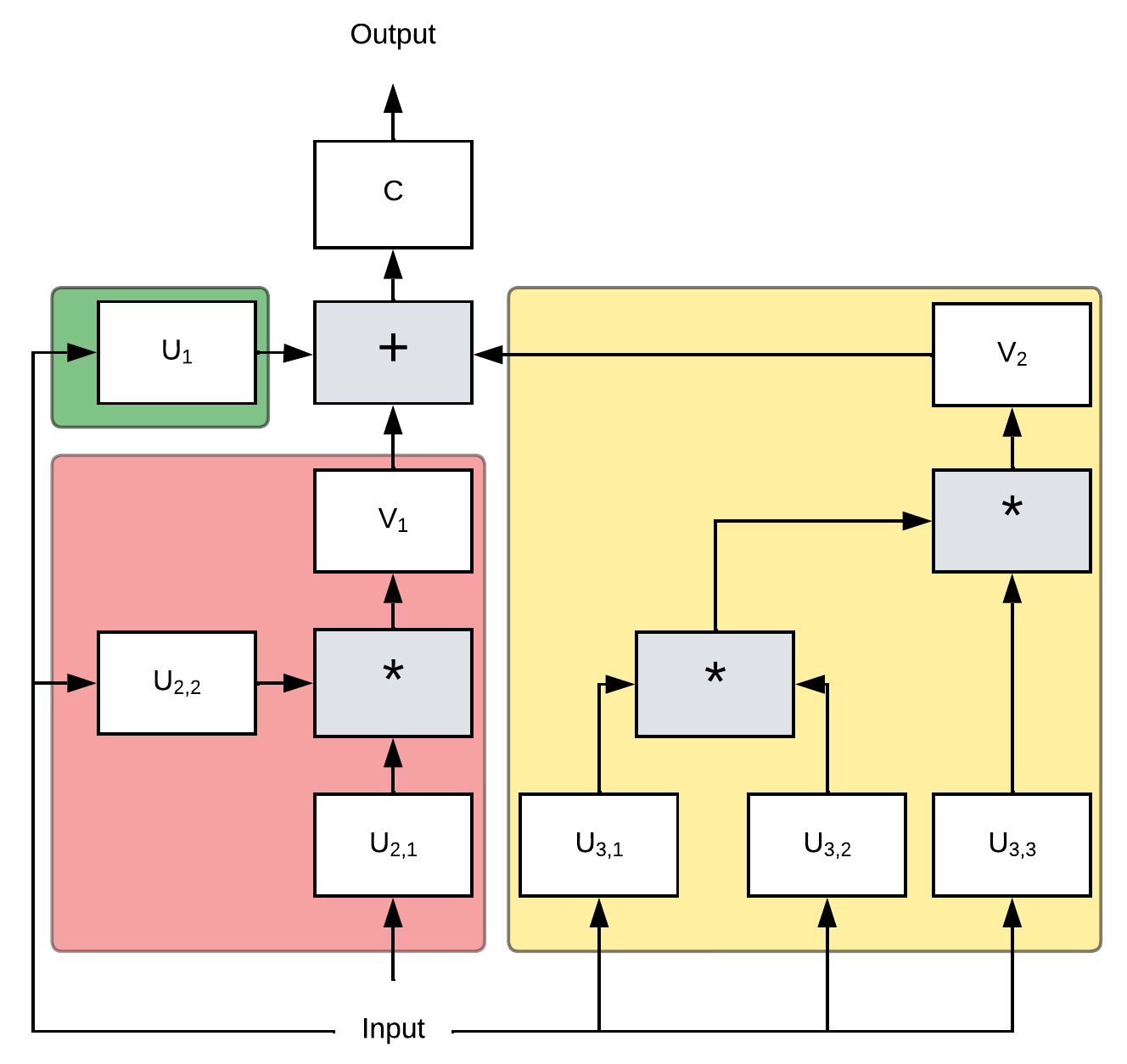}
    \caption{Third-order PDC model \cite{chrysos2022augmenting}.}
    \label{subfig:pdc}
\end{subfigure}
    \hfill
\begin{subfigure}[t]{0.4\textwidth}
    \centering
    \includegraphics[width=\textwidth]{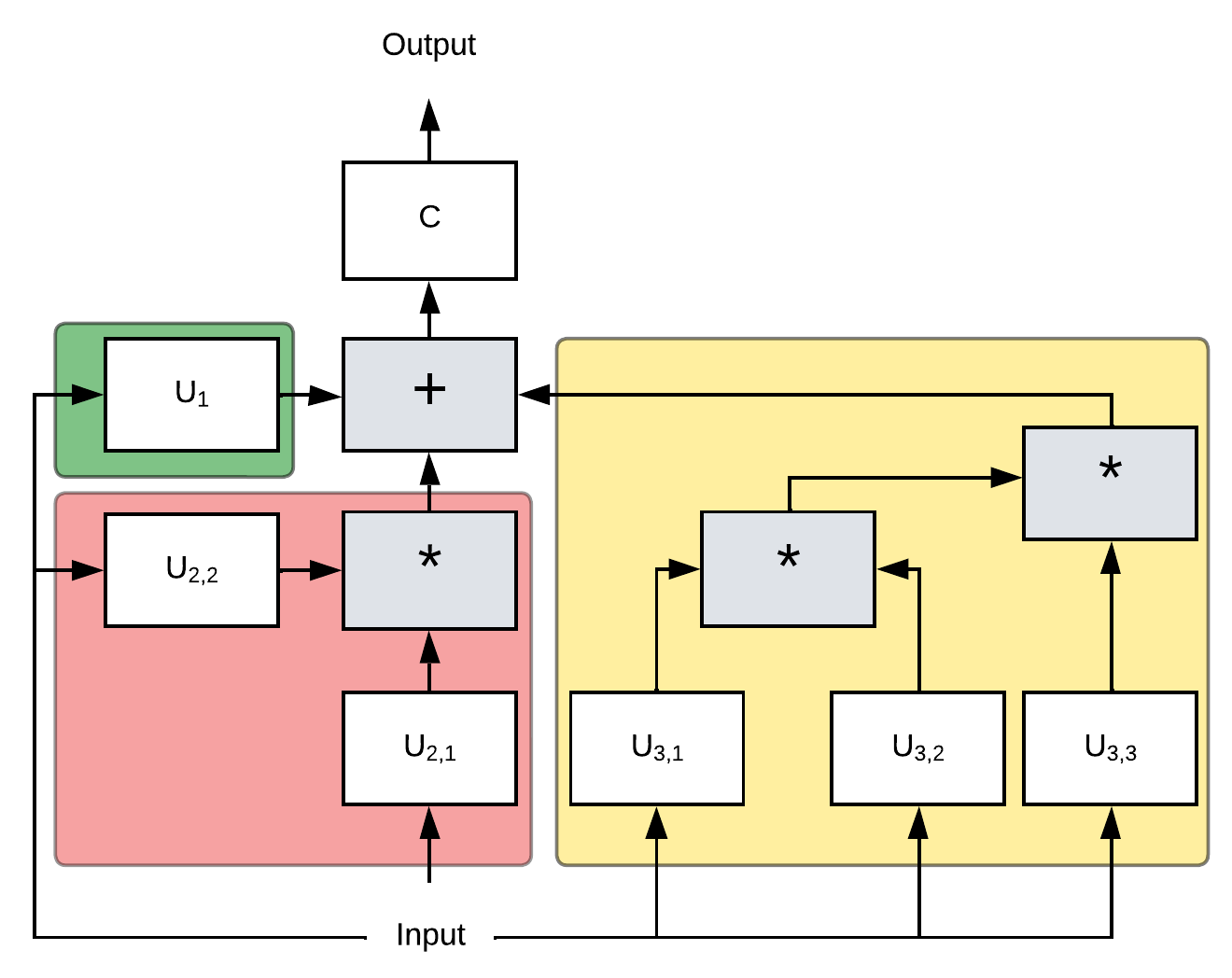}
    \caption{Third-order PDCLow model, obtained from PDC in Figure \ref{subfig:pdc} by removing the $V$ layers.}
    \label{subfig:pdclow}
\end{subfigure}
\hspace{0.05\textwidth}
\caption{Third-order examples of Polynomial Deep Classifier (PDC) models. The architectures can be used recursively for higher-orders than shown here. Our proposal, PDCLow, is essentially a CCP without weight sharing and a low-parameter version of PDC. The colored areas denote the order of the output, where green represents first-order, red is second-order, and yellow is third-order.}
\label{fig:pdcs}
\end{figure}

The work on $\Pi$-Nets is extended with a more expressive model in a separate work, the Polynomial Deep Classifier (PDC) \cite{chrysos2022augmenting}, which also produces a higher-order output of the input. The main difference between PDC and CCP is that weight sharing is no longer assumed across the different orders in the architecture, which allows for higher expressiveness but increases model size. In practice, this means that the special skip connection used in CCP is no longer present. The applications of the two models are similar, although PDC is mainly applied in image and audio classification problems, where different variants of the PDC architecture are embedded in baseline models, which consistently outperform baseline models without PDC embedded. The different PDC variants are achieved by changing the number of channels in the convolutional layers and the depth of the model. In this thesis, the generic PDC model is applied, since the different variants can only be created if convolutional layers are used. A third-order PDC is shown in Figure \ref{subfig:pdc}, where linear layers are shown as white boxes, mathematical operations are shown as gray boxes, and the different colored areas indicate the polynomial order of the output produced by that area, as for the CCP. The absence of weight sharing in PDC is apparent by the increase in the number of $U$ layers, which are multiplied together within each colored area in the figure. For example, raising to second-order is done by multiplying the output of $U_{2,1}$ and $U_{2,2}$ independently of the first-order output from $U_1$. Furthermore, the introduction of hidden layers $V$, significantly increases the depth and number of parameters in PDC compared to CCP. Therefore, to allow comparisons of models without weight sharing and a similar number of parameters, this thesis proposes an architecture in which the $V$ layers are absent, called PDCLow, visualized in Figure \ref{subfig:pdclow}. PDCLow is essentially a CCP without weight sharing and a low-parameter version of PDC.

\subsection{Simulation Metamodeling Approach}\label{sec:sim-meta}
Polynomials are an important tool in simulation, where they are used to relate the values of state variables over time. In discrete event simulations, the state values in the next time step are approximated by polynomials of the previous state values, which have varying order depending on the size of the time step. The size of the time step denotes the time increment from progressing a single step in a simulation. If the size of the time steps is increased, the metamodel approximates state values further ahead in time in the same number of time steps, and the order of the polynomials describing the time step updates increases. Therefore, simulating a given time interval can be done in fewer time steps by modeling higher-order polynomials. Decreasing the number of time steps in a simulation is useful in machine learning, as there are fewer time steps for approximation errors to accumulate. The approach is general to all simulations that use polynomials for time step updates, but the following shows a demonstration for an epidemiology simulation model to better illustrate it.

Epidemiology models represent the spread of an infectious disease in a population over time and consider the population in three different categories or compartments: Susceptible, Infected and Recovered (SIR). Susceptible individuals can contract a disease from infected individuals, whereafter they can recover and are then considered immune to reinfection \cite{downey_2023}. The SIR model in this experiment is a Kermack-McKendrick (KM) model \cite{Kermack-McKendrick}, which models the spread of disease in a closed population. The KM model uses parameters $\beta$ and $\gamma$ to model the rate of infection and recovery, respectively. The infection rate $\beta$ controls the rate of transition from susceptible to infected, whereas the recovery rate $\gamma$ controls the rate from infected to recovered. The transitions between compartments are expressed in terms of the preceding time step, as seen in Equations \ref{eq:sir-s}-\ref{eq:sir-r}.
\begin{align}
    s_{t+1} &= s_{t} - \beta s_{t} i_{t} \label{eq:sir-s} \\ 
    i_{t+1} &= i_{t} + \beta s_{t} i_{t} - \gamma i_{t} \label{eq:sir-i} \\
    r_{t+1} &= r_{t} + \gamma i_{t} \label{eq:sir-r}
\end{align}
The three state variables $s_t$, $i_t$ and $r_t$ describe the fraction of a population $N$ that is susceptible, infected and recovered at a certain time step $t$. The transition equations clarify that there exist two transitions in the KM model, where $\beta s_{t}i_{t}$ determines the fraction of newly infected people in time step $t+1$ and $\gamma i_{t}$ determines the fraction of newly recovered people in time step $t+1$. The state in the time step $t+1$ is fully explained by the state values of the previous time step $t$, which means that the KM model has the memoryless property or the  \emph{Markov property}. Another important property of the KM model for this experiment is that future states can be directly calculated in a recursive manner. For example, if the state of the population in time step $t$ is known, this information can be used to directly calculate the state in time step $t+2$ shown in Equations \ref{eq:sir-st}-\ref{eq:sir-rt}.
\begin{align}
        s_{t+2} &= s_{t+1} - \beta s_{t+1} i_{t+1} \nonumber \\
            &= \beta^3 s_{t}^2 i_{t}^2 - \beta^2 \gamma s_{t} i_{t}^2 + \beta^2 s_{t} i_{t}^2 - \beta s_{t}^2 i_{t} + \beta \gamma s_{t} i_{t} - 2\beta s_{t} i_{t} + s_{t} \label{eq:sir-st}\\
        i_{t+2} &= i_{t+1} + \beta s_{t+1} i_{t+1} - \gamma i_{t+1} \nonumber\\
            &= -\beta^3 s_{t}^2 i_{t}^2 + \beta^2 \gamma s_{t} i_{t}^2 - \beta^2 s_{t} i_{t}^2 + \beta^2 s_{t}^2 i_{t} - 2\beta \gamma s_{t} i_{t} + \beta s_{t} i_{t} + \gamma^2 i_{t} - 2\gamma i_{t} + i_{t} \label{eq:sir-it}\\
        r_{t+2} &= r_{t+1} + \gamma i_{t+1} \nonumber \\
            &= \beta \gamma i_{t} s_{t} - \gamma^2 i_{t} + 2\gamma i_{t} + r_{t} \label{eq:sir-rt}
    \end{align}

These equations are obtained by inserting the definitions from Equations \ref{eq:sir-s}-\ref{eq:sir-r}. When doing so, two time steps are approximated at a time, which means that a lag size of two is used. As the lag size increases, the polynomial order of the transition equations increases. For example, with a lag size of two, the state variable $s$ is described by a seventh-order polynomial as opposed to a third-order polynomial using a lag size of one. In general, as the lag size $L$ increases by one, the polynomial order of the transition equations for $s$ and $i$ increases to $2n_{L-1} + 1$ and for $r$ increases to $2n_{L-1}$, where $n_{L-1}$ is the polynomial order when using a lag size of $L-1$.

\begin{figure}[ht]
    \centering
    \includegraphics[width=\textwidth]{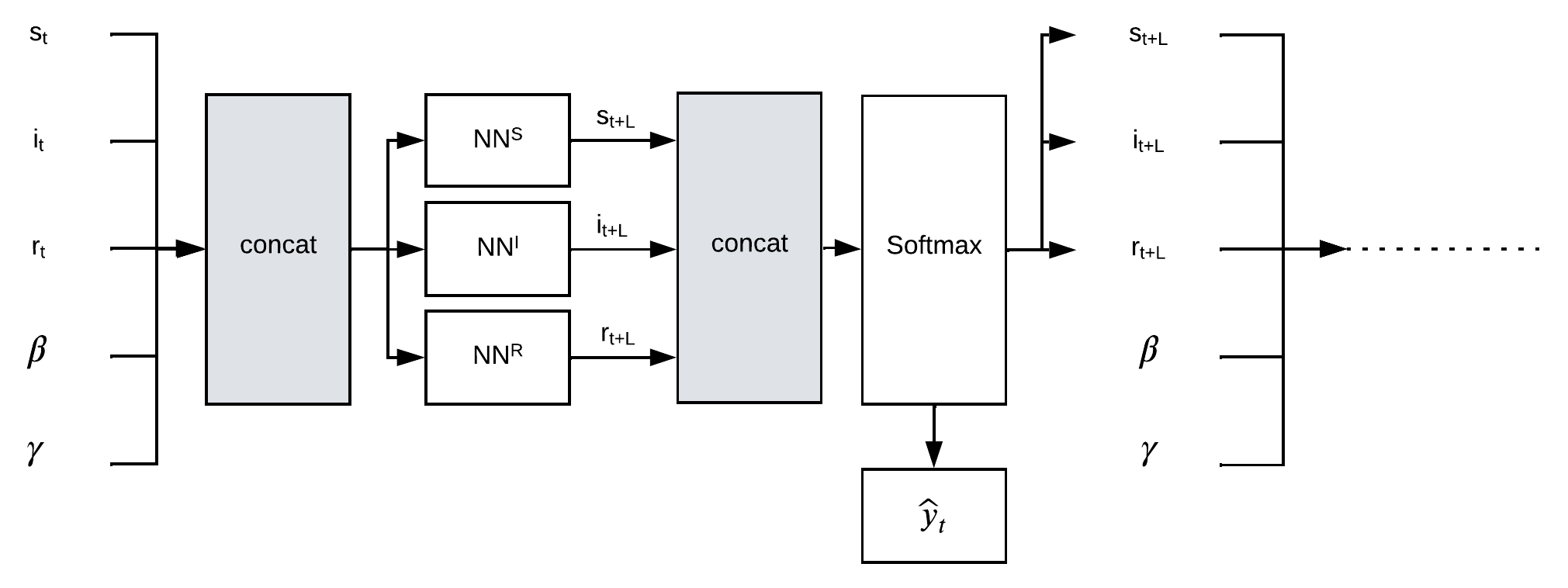}
    \caption{The simulation metamodeling approach and training process of machine learning metamodels. The NN boxes indicate neural networks or other machine learning models and their superscript denotes which compartment they are trying to learn. In simulation with multiple state variables that sum to one, using softmax is useful for regularization of the state value predictions.}
    \label{fig:SIRModel}
\end{figure}

% Write about what happens in each timestep
The learning problem for the metamodel essentially consists of predicting the next state given a previous state. To do this, a model is fed all relevant information about a state, $s_{t}$, $i_t$, and $r_t$, as well as exogenous variables, that are general across all time steps, $\beta$ and $\gamma$. This is illustrated in Figure \ref{fig:SIRModel}, where all the aforementioned values are concatenated into a vector and fed to three neural networks, where the superscripts indicate which compartment they are learning. A neural network is used for each compartment since the influence of the previous state values on the next state values is different for each of them. The influence of the previous state is also largely determined by the lag size, as seen in the transition equations. For example, the transition equation for $r$ with a lag size of one (Equation \ref{eq:sir-r}) depends only on $r$ and $i$ from the previous state. For a lag size of two (Equation \ref{eq:sir-rt}), $r$ depends on all previous state values. The next state predictions  are the concatenated scalar outputs of the neural networks, which also describe fractions of a population that sums to one. This property makes the softmax function useful to ensure that the output has the same property as the input, which is helpful for this specific application in training, validation, and testing. In general, the output of the softmax function is the next state predictions, which are used to calculate the loss along with the true values. However, while validation and testing use the previous state predictions as input for the next state predictions, the true values are used in training. Therefore, errors accumulate for each simulation time step in validation and testing. In addition to the state values, $\beta$ and $\gamma$ are also used to model future time steps, and the above procedure repeats itself throughout the simulation.

\section{Experiments}\label{sec:experiments}
The experiments in this thesis explore the viability of MNNs learning polynomials out-of-distribution and show how MNNs can be used as metamodels for a simulation model with polynomial time step updates. In the first experiment, polynomials are iteratively selected to identify characteristics that make them difficult to learn. The second experiment models synthetic functions from a literature benchmark and compares MNNs with baseline models. The third experiment concerns the use of MNNs as metamodels for an epidemiology simulation model, where transitions between time steps are described by polynomial functions. To carry out these experiments, the following experimental setup is used.

\textbf{Evaluation metrics} are essential for evaluating experimental results, where this thesis uses the root relative square error (RRSE), which is relative to predicting the mean of the true values \cite{RRSE}. This is apparent from Equation \ref{eq:rrse}, where the squared error of a predictor is divided by the squared error of the average predictor. In other words, the RRSE is similar to the root mean square error (RMSE), but scaled by the variance in the output domain. An RRSE of zero is the lowest possible score and means that the true values and predictions match exactly. There is no upper limit to the metric, but an RRSE greater than one indicates that the predictor is worse than an average predictor. Therefore, the relativeness of the RRSE allows performance comparisons between experiments and a perspective in the form of an average predictor. Note that for the RRSE used in the out-of-distribution tests, the average predictor based on the training distribution likely results in an RRSE higher than one on the test set, as the two distributions are different. A description of the additional logged metrics is provided in the Appendix \ref{sec:appendix-metrics}.
\begin{align}
    \text{RRSE} &= \sqrt{\frac{\sum_{i=1}^n (\hat{y}_i-y_i)^2}{\sum_{i=1}^n (y_i- \Bar{y})^2}} \label{eq:rrse} \\
    & \text{where} \, \Bar{y} = \frac{1}{n} \sum_{i=1}^{n} y_i \nonumber
\end{align}

\textbf{Baseline models} are introduced to put the performance of MNNs into perspective. They are simple but powerful machine learning models implemented using the scikit-learn library \cite{scikit-learn}:
\begin{itemize}
    \item Linear Regression (LR)
    \item Gradient Tree Boosting (GB)
    \item Random Forest (RF)
\end{itemize}
Linear regression (LR) is the most simple baseline model, which is fitted by minimizing the residual sum of squares between the targets and predictions from a linear approximation \cite{scikit-learn}. The LR baseline model only considers first-order polynomials, since the feature matrix created by incorporating higher-order interactions between the features can become extremely wide and lead to high computational complexity. In addition to LR, the ensemble methods GB and RF are used. They are both tree-based methods, but are separated by their procedure for creating and aggregating their weak learner trees. GB is a boosting algorithm, where trees are built sequentially based on the previous one. The purpose of building a new tree is to correct the errors of the previous tree and to achieve a strong learner collectively \cite{scikit-learn}. In contrast, RF is a bagging method, which builds its trees in parallel and aggregates its output by averaging its trees \cite{scikit-learn}. This thesis uses the default hyperparameters defined in the scikit-learn documentation \cite{scikit-learn}.

\textbf{Hyperparameters} are determined on the basis of an initial experimental analysis, where the choice of a mean squared error loss function, Adam optimizer \cite{kingma2017adam}, learning rate of 0.001, batch size of 32 and 64 hidden neurons showed good results.

\textbf{Models} are the four MNNs described in Section \ref{subsec:producing-higher-outputs}, which are ideal for generating higher-order output. The number of input layers of CCP, PDC, and PDCLow is determined by the order of the polynomial being learned $n_{order}$, whereas PANN only has one input layer with a polynomial activation function following it. Hidden layers are omitted from PANN because an initial experimental analysis showed sufficient flexibility to learn higher-order polynomials without them. Further tuning of the aforementioned hyperparameters is not conducted due to limitations on computational resources. In summary, the model architectures used are:
\begin{itemize}
    \item \textbf{PANN}: Two-layer FFNN with 64 hidden neurons and $n_{order}$ polynomial activation between the input and output layers. A general example is shown in Figure \ref{fig:pann}.
    \item \textbf{CCP}: Architecture with weight sharing, 64 hidden neurons, and $n_{order}$ input layers. A third-order example is visualized in Figure \ref{fig:ccp}.
    \item \textbf{PDC}: Architecture without weight sharing and with 64 hidden neurons, $\sum_{i=1}^{n_{order}} i$ input layers and $n_{order}-1$ hidden layers. A third-order example is visualized in Figure \ref{subfig:pdc}.
    \item \textbf{PDCLow}: PDC architecture without hidden layers. A third-order example is visualized in Figure \ref{subfig:pdclow}. 
\end{itemize}

Despite using similar mechanics, MNNs have a vastly different number of parameters, depending on the number of input variables, output variables, and the order of the polynomial being learned, which varies greatly in the experiments. Therefore, the expressions to calculate the number of parameters for each MNN are provided below.
\begin{align*}
    \text{PANN parameters} &= n_{i}\cdot n_{h} + n_{h} + n_{h}\cdot n_{o} + n_{o} \\
    \text{CCP parameters} &= n_{order}\left(n_{i}\cdot n_{h} + n_{h}\right) + n_{h}\cdot n_{o} + n_{o} \\
    \text{PDCLow parameters} &= \sum_{d=1}^{n_{order}} \sum_{j=1}^{d} \left( n_{i}\cdot n_{h} + n_{h}\right) + n_{h}\cdot n_{o} + n_{o} \\ 
    \text{PDC parameters} &= \sum_{d=1}^{n_{order}} \left( \sum_{j=1}^{d}( n_{i}\cdot n_{h} + n_{h}) + n_{h}^2 + n_{h} \right) + n_{h}\cdot n_{o} + n_{o}
\end{align*}

The expressions are sorted in ascending order, which means that PANN always has the lowest number of parameters, while PDC always has the highest regardless of the polynomial order. PANN has the same number of layers throughout this thesis, which means that the number of parameters depends only on the input size $n_i$, the number of neurons $n_h$, and the output size $n_o$. For the remaining three models, the number of parameters also depends on the order of the polynomial $n_{order}$. In the special case where a first-order polynomial is modeled, PANN, CCP, and PDCLow have the same number of parameters.

\subsection{Learning Polynomials}
The experiment on polynomial learning is carried out to obtain an indication of the performance of using MNN to learn polynomials in-distribution and generalize out-of-distribution. Polynomials are selected to have additions, subtractions, and multiplications between an arbitrary number of variables of arbitrary degrees. The experiments do not include polynomials with negative exponents or division by variables. The selected polynomials with a varying number of variables, polynomial order, and variable interactions are provided in Table \ref{tab:prelim_exp}. The order of the polynomial is reported as the maximum exponent of the function terms, whereas the degree refers to the exponent of individual terms. Variable interactions are the number of distinct variables multiplied, regardless of their exponent. For example, two variables multiplied together denote two interactions, three variables multiplied are denoted as three interactions, and so on. The polynomials in Table \ref{tab:prelim_exp} are selected in an iterative manner, where the performance of the models learning the "previous" polynomial is taken into account for the next polynomial. The complexity of a polynomial is increased by adding more variables, increasing the order of the polynomial, or the number of interactions. 

\begin{table}[ht]\small
\centering
\caption{A selection of polynomials for experiments with MNNs. Order is the maximum exponent in a polynomial and interactions determines the number of interactions between distinct variables.}
\label{tab:prelim_exp}
\begin{tabular}{|l|l|l|l|}
\hline
\textbf{Polynomial} & \textbf{\# of variables} & \textbf{Order} & \textbf{Interactions} \\ \hline
$a$ & 1 & 1 & 1 \\[0.5ex] \hline
$a^2$ & 1 & 2 & 1\\[0.5ex] \hline
$5a^2 + 6b^2$ & 2 & 2 & 1 \\[0.5ex] \hline
$2a^3 6b^2$ & 2 & 5 & 2 \\[0.5ex] \hline
$2a^3 b^2 - 3c$ & 3 & 5 & 2\\[0.5ex] \hline
$2a^3 b^3 - 3c$ & 3 & 6 & 2\\[0.5ex] \hline
$2a^2 b^2 c^2 - 3d$ & 4 & 6 & 3\\[0.5ex] \hline
$2a^3 b^2 c^2 - d^6$ & 4 & 7 & 3\\[0.5ex] \hline
$a^3 b^2 c^3 - d^3 e^3$ & 5 & 8 & 3\\[0.5ex] \hline
$a^3 b^3 c^3 - d^4 e^4$ & 5 & 9 & 3\\[0.5ex] \hline
$a^3 b^2 c^2 d^3 - e^5 f^5$ & 6 & 10 & 4\\[0.5ex] \hline
$a b c d e f g$ & 7 & 7 & 7 \\[0.5ex] \hline
$a b c + d - e - f - g$ & 7 & 3 & 3 \\[0.5ex] \hline
$a^{10} - b^9$ & 2 & 10 & 1 \\[0.5ex] \hline
\end{tabular}
\end{table}

% Explain about adding noise, using different distribution for training and using relu activation
The experiment is carried out using MNNs to learn each polynomial for 30 epochs with samples from a Gaussian distribution $\mathcal{N}(0,5)$. 
To determine whether a model can generalize out-of-distribution, it is tested on samples from nine different distributions. The distributions are all Gaussian with their mean and standard deviation given by every combination of $\mu=[-50, 0, 90]$ and $\sigma=[1, 5, 25]$. The choices of mean and standard deviation are chosen to ensure nonoverlapping distributions and high numerical diversity in the test sets. An additional small-scale experiment is carried out by training with samples from a Gaussian distribution $\mathcal{N}(0,1)$ to explore the implications of using a different training distribution than $\mathcal{N}(0,5)$.

\subsection{Modeling Synthetic Functions}
In this experiment, synthetic functions are modeled using MNNs and baseline models. This class of functions has previously been used as benchmark functions in engineering problems, such as optimization and metamodeling tasks. The learning problem is to model the input-output functions with the MNNs and the baseline models using samples from their constrained input domains. 

The advantages of using these benchmark functions as opposed to the polynomials of the previous experiment is two-legged. First, by using standard functions, the domain of the input variables is predefined, and the output for this domain does not cause numerical overflow. Second, the complexity of each function is well studied and documented from an engineering perspective. For example, a literature survey comprises the most complete set of global optimization problems \cite{BenchmarkOptimizationProblems}. Although complexity is not reported in the survey, \citet{Gavana} does this through their website on global optimization algorithms. Their work reports not only the performance of global optimization algorithms, but also the share of successful optimizations on particularly tough functions. The subset of functions used in this thesis is selected to include only terms with multiplication(s), addition(s), and subtraction(s) of variables, as in the previous experiment. Furthermore, the functions are selected to have varying polynomial order and complexity in the form of optimization success rates. In addition to the functions of the benchmark set, the functions of Lim \cite{Lim-Polynomial}, Currin \cite{Currin-function}, and Dette-Pepelyshev \cite{DettePepelyshev-Function} are included. An overview of the selected benchmark functions, along with their number of variables, polynomial order, and variable input domain(s) is shown in Table \ref{tab:opt-func}. The functions themselves are left out of the table for brevity, but are found in Appendix \ref{sec:appendix-prelim-exp} (Equations \ref{eq:Currin}-\ref{eq:Goldstein-Price}).

\begin{table}[ht]
\centering
\caption{Benchmark functions used for experiments with MNNs and baseline models. If input domain is described with $x_i$, the specific domain counts for all input variables of the function, i.e. $\forall i \in (1\dots n)$ where $n$ is the number of variables.}
\label{tab:opt-func}
%\resizebox{\textwidth}{!}{%
\begin{tabular}{|l|l|l|l|l|l|}
\hline
\textbf{Author} & \textbf{\# of variables} & \textbf{Polynomial order} & \textbf{Input domain} \\ \hline
Currin \cite{Currin-function} & 2 & 2 & $x_i \in (0,1)$ \\[0.5ex] \hline
Bukin 6 \cite{Bukin-Function} & 2 & 2 & $x_1 \in (-15, 5), \, x_2 \in (-3,3)$ \\[0.5ex] \hline
Price 3 \cite{Price3-Function} & 2 & 4 & $x_i \in (-500, 500)$ \\[0.5ex] \hline
Dette-Pepelyshev \cite{DettePepelyshev-Function} & 3 & 4 & $x_i \in (0,1)$\\[0.5ex] \hline
Colville \cite{BenchmarkOptimizationProblems} & 4 & 4 & $x_i \in (-10, 10)$ \\[0.5ex] \hline
Lim \cite{Lim-Polynomial} & 2 & 5 & $x_i \in (0,1)$ \\[0.5ex] \hline
Camel Three Hump \cite{Camel-Function} & 2 & 6 & $x_i \in (-5,5)$ \\[0.5ex] \hline
Beale \cite{BenchmarkOptimizationProblems} & 2 & 8 & $x_i \in (-4.5,4.5)$ \\[0.5ex] \hline
Goldstein-Price\cite{GoldsteinPrice-Function} & 2 & 8 & $x_i \in (-2,2)$ \\[0.5ex] \hline
\end{tabular}
%}

\end{table}

The experiment is carried out by learning the functions in Table \ref{tab:opt-func} over 100 epochs with the four MNNs and the three baseline models. The data is generated by drawing 100,000 samples for each variable from a uniform distribution with minimum and maximum according to their input domain, sorting the samples, and using the 5-95 percentile range for training and the remaining for testing. Outside the 5-95 percentile range, the input-output behavior of the synthetic functions is often significantly different from that inside the range, which can be viewed in Appendix \ref{sec:appendix-opt-func}. The different input-output behavior in training compared to testing serves the same purpose as testing with samples from a different distribution. Therefore, this is still considered an out-of-distribution test.

\subsection{Epidemiology Simulation Model}
The third experiment concerns the task of MNNs being metamodels for the SIR simulation model, which is described along with the metamodeling approach in Section \ref{sec:sim-meta}, while this section focuses on how the experiment and the out-of-distribution tests are carried out. In this experiment, the metamodels are trained with SIR simulation data generated with a set of parameters, where the aim is to be able to generalize out-of-distribution when different parameters are used. The parameters of the SIR simulation are considered to be the starting state consisting of $s_0$, $i_0$, $r_0$, the infection rate $\beta$, the recovery rate $\gamma$, the duration of the simulation $T$, and the lag size $L$. 

% Starting state
The starting state values $s_0$, $i_0$, and $r_0$ represent the fraction of the population that is susceptible, infected, or recovered in the first time step of the simulation. Therefore, the starting state values are between zero and one, summing to one within a single time step. A set of starting state values is randomly generated for each simulation run by sampling values from $\text{Uniform}(0, 100)$ and dividing each of them by their sum, which is considered the population size. The division by population size ensures that fractions of the population are modeled instead of absolute counts. This is mandatory, as transition equations described in Section \ref{sec:sim-meta} would end up modeling negative counts of people otherwise. 

\begin{figure}[ht]
    \centering
    \hspace{0.075\textwidth}
    \begin{subfigure}[t]{0.35\textwidth}
    \centering
    \includegraphics[width=\textwidth]{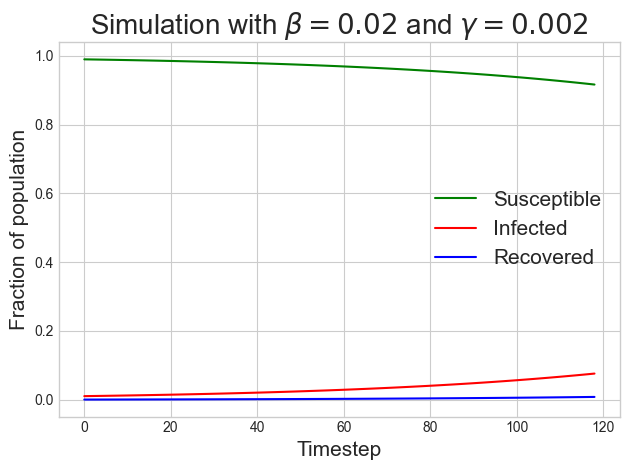}
    \caption{SIR simulation using the lower bound training values for $\beta$ and $\gamma$.}
    \label{subfig:SIR_train_min}
    \end{subfigure}
    \hfill
    \begin{subfigure}[t]{0.35\textwidth}
    \centering
    \includegraphics[width=\textwidth]{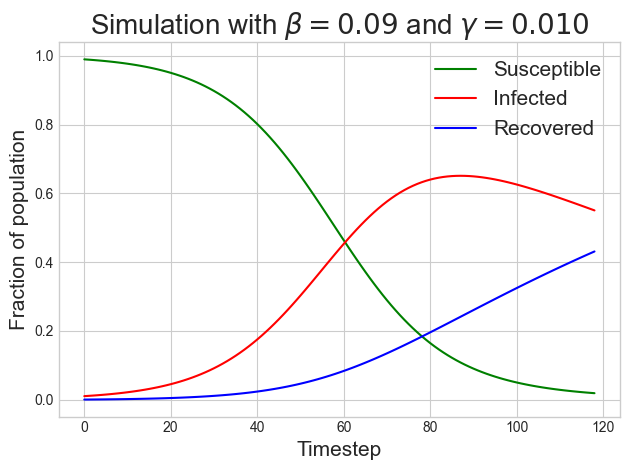}
    \caption{SIR simulation using the upper bound training values for $\beta$ and $\gamma$.}
    \label{subfig:SIR_train_max}
    \end{subfigure}
    \hspace{0.075\textwidth}

    \hspace{0.075\textwidth}
    \begin{subfigure}[t]{0.35\textwidth}
    \centering
    \includegraphics[width=\textwidth]{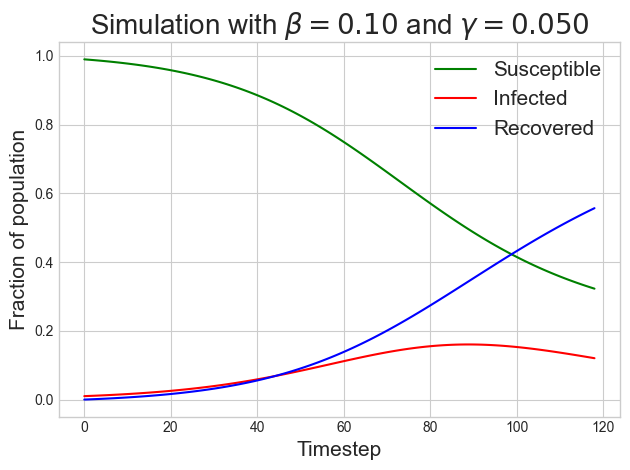}
    \caption{SIR simulation using the lower bound testing values for $\beta$ and $\gamma$.}
    \label{subfig:SIR_test_min}
    \end{subfigure}
    \hfill
    \begin{subfigure}[t]{0.35\textwidth}
    \centering
    \includegraphics[width=\textwidth]{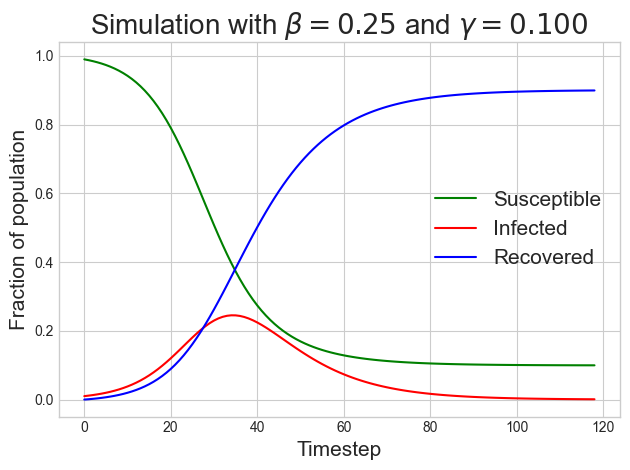}
    \caption{SIR simulation using the upper bound testing values for $\beta$ and $\gamma$.}
    \label{subfig:SIR_test_max}
    \end{subfigure}
    \hspace{0.075\textwidth}
    \caption{SIR simulations with the same starting state of $s_0=0.99, i_0=0.01$ and $r_0=0.$ and varying $\beta$ and $\gamma$ values. The graphs in the upper row use the training parameter values, and the bottom row uses the testing parameter values.}
    \label{fig:epidemiology_train_test_visualization}
\end{figure}

% Beta and gamma
In relation to the transition equations, the infection rate $\beta$ and the recovery rate $\gamma$ control the transition rate between the three compartments and are sampled in each simulation run like the starting state values. Likewise, $\beta$ is sampled from $\text{Uniform}(0.02, 0.09)$ and $\gamma$ is sampled from $\text{Uniform}(0.022, 0.01)$ in training, while being sampled from $\text{Uniform}(0.1, 0.25)$ and $\text{Uniform}(0.05, 0.1)$ in test, respectively. The two distributions for each parameter are nonoverlapping, implying that models are tested with out-of-distribution $\beta$ and $\gamma$ values. Simulations with the lower and upper limits of the values of $\beta$ and $\gamma$ in training and testing are visualized in Figure \ref{fig:epidemiology_train_test_visualization}. The four subplots in the figure show the evolution of the three compartments, Susceptible, Infected, and Recovered, over a duration of 120 time steps from the same starting state. The first simulation in Figure \ref{subfig:SIR_train_min} shows a simulation with very low values of $\beta$ and $\gamma$. Thus, the rate of transition between the three compartments is slow, as seen by the small change in the simulation from its starting state. The second simulation in Figure \ref{subfig:SIR_train_max} shows a rapid decrease for the susceptible, a rapid increase for the infected, and a slow increase for the recovered. This indicates that the value of $\beta$ is high compared to the value of $\gamma$. The third simulation in Figure \ref{subfig:SIR_test_min} shows a slow decrease for the susceptible and a slow increase for the infected and recovered. The infected decreases toward the end of the simulation, which could indicate that the value of $\gamma$ is relatively high. Finally, the fourth simulation in Figure \ref{subfig:SIR_test_max} shows high values of $\beta$ and $\gamma$ seen by a sharp decrease in susceptible and a sharp increase in recovered. These simulations show that varying $\beta$ and $\gamma$ values can lead to different transition behaviors. Therefore, models are required to learn a range of input-output behaviors in this experiment.

% Duration and lag size
The final parameters ($T$ and $L$) are essential to test the viability of MNNs in metamodeling. The duration of the simulation $T$ defines the time interval, which the simulation spans over, while the lag size $L$ describes the size of the time steps. These parameters are closely related, as the number of modeled time steps decreases if the lag size is increased for a fixed duration. To be exact, the number of time steps modeled for a simulation is given by $T//L$, where $//$ is the integer division operator. The number of modeled time steps must be above one for the learning problem to be meaningful, which entails that the duration and lag size must be chosen such that $T//L>1$. This experiment is carried out by varying the duration of the simulation $T$ and the lag size $L$ to explore its implications on the performance of MNNs and baseline models. First, the impact of the simulation duration is tested for the values $T=[2, 6, 12, 24, 30, 40, 60, 120]$ with a fixed lag size of $L=1$. For this lag size, an MNN is constructed with a polynomial order of three for each of the three compartments. In terms of the duration, it is expected that performance worsens as the duration increases, as there are more time steps for errors to accumulate. For the second part of this experiment, the duration is fixed at $T=120$ and the lag size is tested for the values $L=[1, 2, 3, 4, 5]$. Increasing the lag size decreases the number of modeled time steps in a simulation, as explained by the expression $T//L$. Consequently, the number of modeled time steps for these lag sizes and a duration of 120 are found to be $T//L=[120, 60, 40, 30, 24]$, respectively. Therefore, the tested values of duration and lag size are chosen in a way that allows comparison of runs with the same number of modeled time steps but with different lag sizes, and thereby different polynomial order. Conducting this experiment involves modeling polynomials of orders 3, 7, 15, 31, and 63, respectively. The downside to modeling such high-order polynomials is that the number of model parameters increases drastically, as seen in Table \ref{tab:model-params}, where the models are sorted in increasing order by their number of parameters.

\begin{table}[ht]
\centering
\caption{Model parameters for when using MNN building blocks in SIR simulation model for different lag sizes $L$. The numbers in parentheses denote the polynomial order incurred by the corresponding lag size.}
\label{tab:model-params}
\begin{tabular}{|l|l|l|l|l|l|}
\hline
\multicolumn{1}{|c|}{\textbf{Building Block}} & \multicolumn{1}{c|}{\textbf{\begin{tabular}[c]{@{}c@{}}$L=1$\\ (3)\end{tabular}}} & \multicolumn{1}{c|}{\textbf{\begin{tabular}[c]{@{}c@{}}$L=2$\\ (7)\end{tabular}}} & \multicolumn{1}{c|}{\textbf{\begin{tabular}[c]{@{}c@{}}$L=3$\\ (15)\end{tabular}}} & \multicolumn{1}{c|}{\textbf{\begin{tabular}[c]{@{}c@{}}$L=4$\\ (31)\end{tabular}}} & \multicolumn{1}{c|}{\textbf{\begin{tabular}[c]{@{}c@{}}$L=5$\\ (63)\end{tabular}}} \\ \hline
PANN & 1.3K & 1.3K & 1.3K & 1.3K & 1.3K \\\hline
CCP & 3.6K & 8.2K & 17.4K & 35.9K & 72.7K \\ \hline
PDCLow & 7.1K & 32.4K & 138K & 571K & 2.3M \\ \hline
PDC & 44.5K & 119K & 325K & 958K & 3.1M \\ \hline
\end{tabular}
\end{table}

For each experimental run, 100,000 simulations are generated and a batch size of 32 is used. As a result, the feature matrix has dimension $[B, T//L-1, F] = [32, T//L-1, 5]$, since the same batch size $B$ and the same number of features $F$ are used for all runs. The features include the three state variables and the two transition rates $\beta$ and $\gamma$. The middle dimension of the feature matrix is subtracted by one as the objective is to predict the next state from a previous one. Naturally, this means that the target matrix has a similar dimension $[B, T//L-1, F_{state}] = [32, T//L-1, 3]$, where $F_{state}$ denotes the three state variables in the simulation.

\section{Results}\label{sec:results}
The results of the three experiments described in the previous section are presented and interpreted in this section. The first experiment aims to identify the characteristics of polynomials that make it difficult to learn in-distribution and generalize out-of-distribution. The second experiment consists of modeling a set of synthetic functions with MNNs and baseline models, which further tests the viability of MNNs learning polynomial functions out-of-distribution. The third experiment uses MNNs and baseline models as metamodels for an epidemiology simulation model, where duration of the simulation and lag size are varied. 

\subsection{Polynomials}
\begin{figure}[ht]
    \centering
    \hfill
    \begin{subfigure}[t]{0.23\textwidth}
    \centering
    \includegraphics[width=\textwidth]{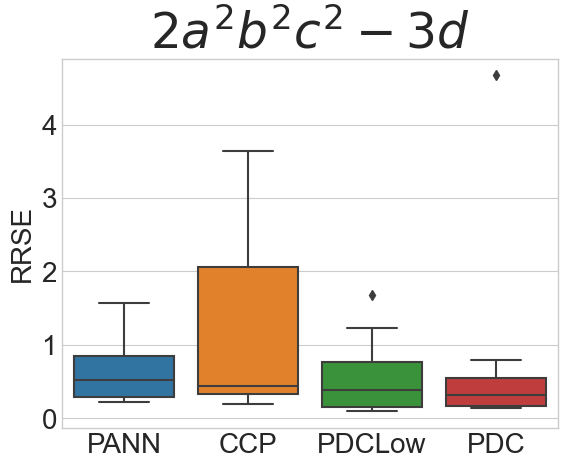}
    \caption{Sixth-order polynomial with four variables.}
    \label{subfig:2a2_b2_c2-3d}
    \end{subfigure}
    \hfill
    \begin{subfigure}[t]{0.23\textwidth}
    \centering
    \includegraphics[width=\textwidth]{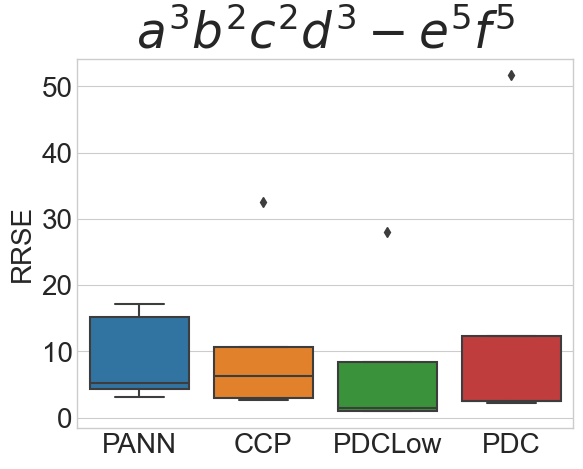}
    \caption{Tenth-order polynomial with six variables.}
    \label{subfig:a3_b2_c2_d3-e5_f5}
    \end{subfigure}
    \hfill
    \begin{subfigure}[t]{0.23\textwidth}
    \centering
    \includegraphics[width=\textwidth]{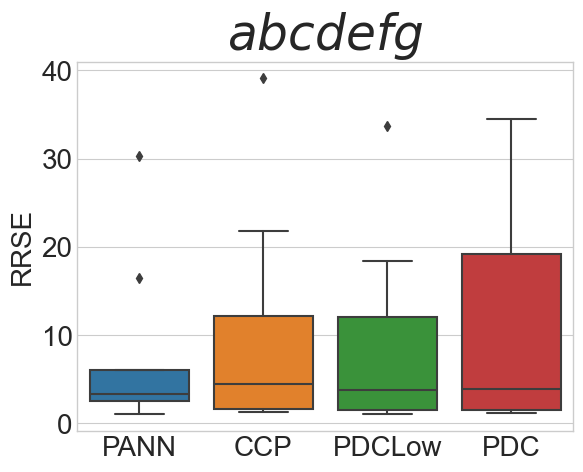}
    \caption{Seventh-order polynomial with seven variables.}
    \label{subfig:a_b_c_d_e_f_g}
    \end{subfigure}
    \hfill
    \begin{subfigure}[t]{0.23\textwidth}
    \centering
    \includegraphics[width=\textwidth]{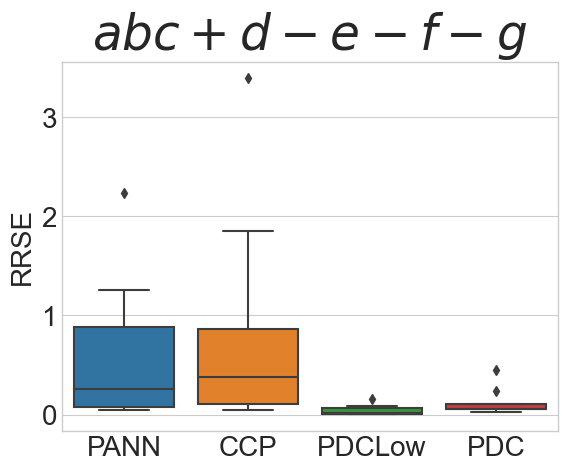}
    \caption{Third-order polynomial with seven variables.}
    \label{subfig:a_b_c+d-e-f-g}
    \end{subfigure}
    \hfill
    \caption{Performance of MNNs - PANN, CCP, PDCLow, and PDC - for four different polynomials from Table \ref{tab:prelim_exp}. Values for each box-plot are the RRSE scores when tested with the nine test distributions. The whiskers extend to $1.5\times \text{IQR}$ and the outliers are marked by diamonds.}
    \label{fig:polynomial_boxplots}
\end{figure}

The polynomials in Table \ref{tab:prelim_exp} are ordered in terms of the iterative process in which they were chosen. As a result, the complexity of the polynomials gradually increases until performance starts to decrease drastically. Up until the sixth-order, the performance is great in validation and remains satisfactory in the tests. Beyond the sixth-order, MNNs perform worse than an average predictor in validation, and the results in the tests are even worse, as seen in the Appendix \ref{sec:appendix-prelim-exp}. To avoid redundancy in this section, evaluation is done for four selected polynomials shown in Figure \ref{fig:polynomial_boxplots}, where performance is visualized as box plots of the RRSE scores in the test distributions. The colored boxes denote the interquartile range (IQR) of the RRSE test scores for each MNN in the nine test distributions. Moreover, the whiskers extend to $1.5\times \text{IQR}$ and all scores beyond this are considered outliers and marked with diamonds.

Figure \ref{subfig:2a2_b2_c2-3d} shows the RRSE test scores for the polynomial $2a^2 b^2 c^2 - 3d$, which is a sixth-order polynomial with four variables. The IQR of the models are located between zero and one, with the exception of CCP, whose RRSE score is high in the test distributions $\mathcal{N}(-50,1)$, $\mathcal{N}(0,1)$, and $\mathcal{N}(90,1)$ compared to the rest of the test distributions. This is the case for all models, but the difference in RRSE is less significant than for CCP. In general, PDC has the lowest RRSE scores, indicated by the low placement of the IQR, but it has an outlier with an RRSE score greater than four, which is recorded in the Gaussian $\mathcal{N}(0,1)$ distribution.  

The results of a more complex polynomial are shown in Figure \ref{subfig:a3_b2_c2_d3-e5_f5}, where the RRSE scores have increased drastically, as seen by the range of the y-axis. As the polynomial order increases, the chance of numerical overflow and producing a "not a number" (NaN) increases as well. In this case with a tenth-order polynomial, only five of the nine test distributions are usable, as the rest produce NaNs. Therefore, the performance is likely worse than what the figure shows, which is already terrible. However, PDCLow performs well in most test distributions, with the exception of a worse performance in $\mathcal{N}(0,5)$ and a terrible performance in $\mathcal{N}(0,1)$. PDCLow and CCP have much greater overall performance than the much larger PDC model for the complex, tenth-order polynomial, whereas PDC has better performance than PDCLow and CCP in the less complex, sixth-order polynomial previously seen.

As NaNs are encountered for the tenth-order polynomial, the order of the polynomial decreases and the number of variables increases. The performance in Figure \ref{subfig:a_b_c_d_e_f_g} shows that the RRSE has similar magnitude as the previous polynomial, but longer whiskers and more outliers are present. This is potentially attributed to the fact that none of the test distributions produce NaNs and that the performance is more true to reality than previously seen. In terms of their RRSE scores, it is difficult to separate the models. The median and lower quartile are very close for CCP, PDCLow, and PDC, whereas the median for PANN is slightly lower, but the lower quartile is higher than the three other models. PANN also has a surprisingly narrow IQR, which indicates that it performs consistently for this polynomial. This is likely a result of the PANN architecture, which only uses multiplication to generate higher-order output, where the remaining MNNs use both multiplication and addition, which preserves modeling of lower degree terms. As this polynomial contains only multiplication, PANN can have a stronger inductive bias towards learning polynomials of this type. Outliers for this polynomial are produced by the test distributions $\mathcal{N}(-50,1)$ and $\mathcal{N}(90,1)$, but the performance is generally unsatisfactory.

In an attempt to get around this, the number of variables remains fixed, but the polynomial order is significantly decreased. Figure \ref{subfig:a_b_c+d-e-f-g} shows the RRSE scores of the MNNs for a seven-variable third-order polynomial, where the magnitude of the y-axis has decreased greatly. It is evident that PDC and PDCLow are superior to CCP and PANN, as seen by their narrow IQR close to zero, which is a perfect performance in RRSE. The outliers of PDC and PDCLow are below an RRSE of 0.5, which is very low compared to the RRSE outliers of CCP and PANN recorded on the $\mathcal{N}(90,1)$ test distribution.

% Talk about number of variables and orders we are able to fit
% Talk about distributions. Which ones are we not generalizing to and why?
% Talk about models. Is higher model complexity equal to better performance? Address outputting NaNs
% Did adding noise work?
% Did another training distribution do anything to performance?
% Did ReLU improve performance?
From the results presented above and shown in Appendix \ref{sec:appendix-prelim-exp}, some patterns in terms of MNN performance in learning and generalizing polynomials are becoming apparent. The number of variables alone has no significant impact on the ability to learn a polynomial out-of-distribution. The order of the polynomial seems to have an effect, as models struggled for polynomials above sixth-order. A possible cause is that MNN models, with the exception of PANN, grow larger as the polynomial order increases, which requires them to be trained for longer before learning starts to converge. However, this has not been pursued in this experiment due to limited computational resources. Furthermore, in higher-order polynomials, small changes in the input variables can lead to outputs numerically far from each other, which is more difficult to learn for neural networks. Another issue with higher-order polynomials is the increasing chance of numerical overflow as seen for some of the test distributions in this experiment. While the polynomial order has a clear impact on out-of-distribution performance, it is difficult to identify the impact of the number of interactions between variables, since it is closely related to the polynomial order. In terms of testing distributions, it is apparent that distributions with a standard deviation of one caused many outliers of RRSE. This could be explained by the properties of the training distribution $\mathcal{N}(0,5)$, which is a wider distribution than the test distributions with a standard deviation of one. The additional small-scale experiment showed that when training is conducted with a Gaussian $\mathcal{N}(0,1)$, outliers are present in different test distributions than training with a Gaussian $\mathcal{N}(0,5)$. While this is an interesting problem to explore further, it is not pursued in this thesis.

\subsection{Synthetic Functions}
% write how experiment with optimization functions addresses some concerns of previous experiment
In the previous experiment with polynomials, numerical overflow is encountered in the test distributions of tenth-order polynomials. Additionally, for these polynomials, small changes in the input lead to large output ranges, which are harder for MNNs to learn and generalize to. These issues are addressed in this experiment by using synthetic functions, where constraints are placed on the domain of the input variables. In most of the functions, this entails function output that is also constrained to a reasonable range. The results of using MNNs and baseline models to learn the synthetic functions of Table \ref{tab:opt-func} are shown in Table \ref{tab:opt-func-results}. The table shows the RRSE in testing for the models in each function, where the first four columns are the results of the baseline models and the last four columns are for the MNNs. The numbers in bold indicate which model has the lowest RRSE for a specific synthetic function.

\begin{table}[ht]
\centering
\caption{Out-of-distribution testing RRSE of baseline and MNN models for all the synthetic functions. Bold indicates the lowest RRSE score for a specific function.}
\label{tab:opt-func-results}
\resizebox{\textwidth}{!}{%
\begin{tabular}{|l|l|l|l|l|l|l|l|l|}
\hline
 & \multicolumn{4}{c|}{\textbf{Baseline}} & \multicolumn{4}{c|}{\textbf{MNN}} \\ \hline
\textbf{Author} & \textbf{Avg.} & \textbf{LR} & \textbf{GB} & \textbf{RF} & \textbf{PANN} & \textbf{CCP} &\textbf{PDCLow} & \textbf{PDC} \\ \hline
Currin \cite{Currin-function} & 1.282 & 8.78e-01 & 1.89e-01 & 1.61e-01 & 4.52e-06 & 1.06e-05 & 4.25e-06 & \textbf{4.11e-07} \\ \hline
Bukin 6 \cite{Bukin-Function} & 1.301 & 1.189 & 0.158 & \textbf{0.121} & 0.660 & 0.660 & 0.667 & 0.652 \\ \hline
Price 3 \cite{Price3-Function} & 9.162 & 9.067 & 2.620 & \textbf{2.611} & 11.211 & 11.132 & 11.124 & 11.120 \\\hline
Dette-Pepelyshev \cite{DettePepelyshev-Function} & 3.184 & 3.109 & 0.802 & 0.815 & 0.208 & 0.040 & 0.412 & \textbf{0.031} \\ \hline
Colville \cite{BenchmarkOptimizationProblems} & 5.044 & 5.021 & 1.685 & 1.613 & 0.213 & 0.114 & 0.036 & \textbf{0.020} \\ \hline
Lim \cite{Lim-Polynomial} & 1.002 & 0.241 & 0.138 & 0.138 & 0.033 & 0.018 & 0.018 & \textbf{0.009} \\ \hline
Camel Three Hump \cite{Camel-Function} & 4.695 & 4.686 & 1.891 & 1.878 & 0.168 & \textbf{0.040} & 0.126 & 0.055 \\ \hline
Beale \cite{BenchmarkOptimizationProblems} & 5.170 & 5.227 & 2.388 & 2.284 & 0.064 & 0.074 & 0.023 & \textbf{0.022} \\ \hline
Goldstein-Price \cite{GoldsteinPrice-Function} & 1.271 & 1.168 & 0.310 & 0.299 & \textbf{0.102} & 0.150 & 0.134 & 0.107 \\ \hline
\end{tabular}
}
\end{table}

% PDC is the best model overall, followed by RF baseline, and PANN and CCP
% No correlation between polynomial order and RRSE score or the complexity as reported by Gavana
% Why is Price so high? Talk about large range of the function output
In addition to the MNNs and baseline models already introduced, the RRSE of the average predictor based on the training data (Avg. in Table \ref{tab:opt-func-results}) is also shown to highlight the severity of the out-of-distribution test and the fluctuations of each function. When MNNs are tested on this set of synthetic functions, the PDC has the lowest RRSE for five of the nine functions, while PANN and CCP have the lowest RRSE for one function each, both having a polynomial order greater than six. PDCLow falls closely behind the best performing models, but does not have superior performance for any function.
Interestingly, the baseline RF model has the lowest RRSE for two of the functions, both of which have a relatively low polynomial order of two and four. From these results, there is no indication that higher-order polynomials are harder to learn than lower-order polynomials. For example, low RRSE scores of the MNNs are seen for the eight-order Beale function, whereas significantly higher RRSE values are seen for the second-order Bukin function. While the Bukin function is reported as the toughest function and the Beale function is reported as the simplest in terms of optimization success rate, there are no other similar relationships between the optimization success rate and the test RRSE. Furthermore, high RRSE scores are reported for all models in the Price 3 function, where none of the MNNs are better than an average predictor based on the training data. The consistently high RRSE for this function could be due to the wide range of function output, which ranges from 0 to ${2}\mathrm{e}{+13}$. In comparison, the next-widest output range in the Beale function ranges from 0 to ${1.8}\mathrm{e}{+5}$, where most models have low RRSE scores.

\begin{table}[ht]
\centering
\caption{In-distribution validation RRSE of baseline and MNN models for all the synthetic functions. Bold indicates the lowest RRSE score for a specific function. The average predictor is omitted from this table as it is has an RRSE of one for all functions.}
\label{tab:opt-func-val}
\resizebox{\textwidth}{!}{%
\begin{tabular}{|l|l|l|l|l|l|l|l|}
\hline
 & \multicolumn{3}{c|}{\textbf{Baseline}} & \multicolumn{4}{c|}{\textbf{MNN}} \\ \hline
\textbf{Author} & \textbf{LR} & \textbf{GB} & \textbf{RF} & \textbf{PANN} & \textbf{CCP} &\textbf{PDCLow} & \textbf{PDC} \\ \hline
Currin \cite{Currin-function} & 5.04e-01 & 3.67e-02 & 3.85e-03 & 3.15e-06 & 6.61e-06 & 3.67e-06 & \textbf{5.35e-07} \\ \hline
Bukin 6 \cite{Bukin-Function} & 0.702 & 0.136 & \textbf{0.009} & 0.488 & 0.496 & 0.489 & 0.486 \\ \hline
Price 3 \cite{Price3-Function} & 1.000 & 0.026 & \textbf{0.007} & 1.424 & 1.415 & 1.418 & 1.417 \\\hline
Dette-Pepelyshev \cite{DettePepelyshev-Function} & 0.919 & 0.086 & 0.026 & 0.090 & 0.018 & 0.109 & \textbf{0.009} \\ \hline
Colville \cite{BenchmarkOptimizationProblems} & 0.986 & 0.068 & 0.050 & 0.062 & 0.021 & 0.015 & \textbf{0.006} \\ \hline
Lim \cite{Lim-Polynomial} & 0.620 & 0.029 & \textbf{0.004} & 0.032 & 0.009 & 0.009 & 0.005 \\ \hline
Camel Three Hump \cite{Camel-Function} & 1.000 & 0.016 & \textbf{0.003} & 0.038 & 0.005 & 0.015 & 0.007 \\ \hline
Beale \cite{BenchmarkOptimizationProblems} & 0.999 & 0.113 & 0.016 & 0.038 & 0.015 & 0.006 & \textbf{0.005} \\ \hline
Goldstein-Price \cite{GoldsteinPrice-Function} & 0.861 & 0.065 & \textbf{0.007} & 0.120 & 0.089 & 0.106 & 0.086 \\ \hline
\end{tabular}
}
\end{table}

% Test - validation comparison: MNNs way better in testing but less better in validation
% MNNs validation performance is true to their test performance
MNNs generally have considerably lower RRSE scores than baseline models in testing. However, in validation, baseline models often outperform MNNs in terms of RRSE, as seen in Table \ref{tab:opt-func-val}. From validation to testing, the RRSE of the baseline models increases signficantly, whereas the RRSE scores for MNNs in validation are much closer to the RRSE scores in testing. Therefore, the validation performance of MNNs is more representative of the out-of-distribution testing performance than it is for the baseline models. The fact that baseline models struggle in testing and that the average predictor based on the training data generally has much higher RRSE than one, indicates that testing is out-of-distribution and highlights the inductive biases of MNNs towards generalizing polynomials.

\subsection{Epidemiology Simulation Metamodels}
In both of the previous experiments, it is found that polynomial functions are harder to learn when their output range is extremely wide. Additionally, it is found that increasing the order of polynomials does not lead to increasing complexity by itself. The first finding is addressed in this experiment by having constrained input and output ranges, and the latter is further investigated by modeling polynomials up to 63rd order. The results of varying the duration of the simulation by $T=[2, 6, 12, 24, 30, 40, 60, 120]$ and keeping the lag size fixed at $L=1$ are the first to be presented in this section. The next results presented are achieved by keeping the duration fixed at $T=120$ and varying the lag size by $L=[1, 2, 3, 4, 5]$. The final results of this experiment come from comparing simulations with the same number of modeled time steps ($T//L$) but with a different lag size and, therefore, with a different polynomial order.

\begin{figure}[ht]
    \centering
    \hspace{0.05\textwidth}
    \begin{subfigure}[t]{0.4\textwidth}
    \centering
    \includegraphics[width=\textwidth]{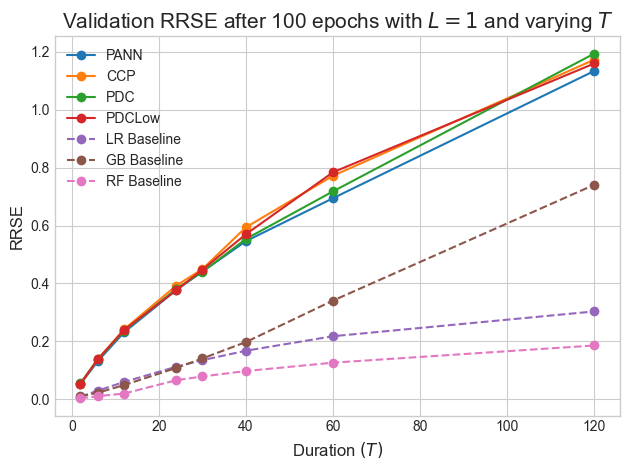}
    \caption{Validation performance where MNNs are outperformed by baseline models.}
    \label{subfig:SIR_val_fixed_l}
    \end{subfigure}
    \hfill
    \begin{subfigure}[t]{0.4\textwidth}
    \centering
    \includegraphics[width=\textwidth]{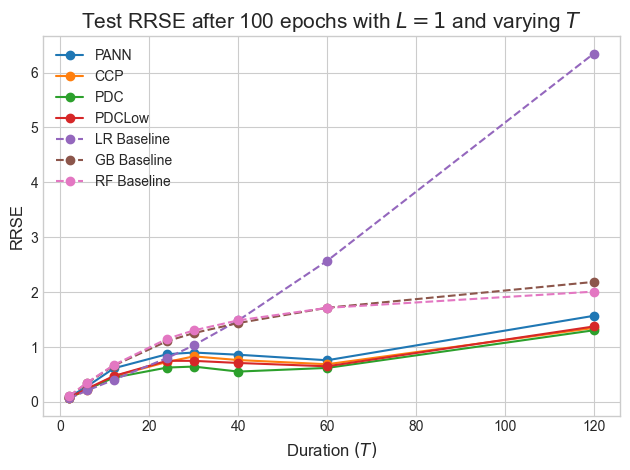}
    \caption{Testing performance where MNNs outperform baseline models.}
    \label{subfig:SIR_test_fixed_l}
    \end{subfigure}
    \hspace{0.05\textwidth}
\caption{Experiment with fixed lag size of one and varying length of simulation. Baseline models are shown with dotted lines and MNNs are shown with solid lines. The y-axis shows the RRSE score, which means lower scores are better.}
\label{fig:SIR_sequence_results}
\end{figure}

% Baselines outperform MNNs in validation
    % RF is consistently the best model
% MNNs outperforms baselines in testing
    % LR is the best model for first three simulation lengths
% Interpret RRSE - above one is worse than average predictor, or is it?
The first results are shown in Figure \ref{fig:SIR_sequence_results}, where two graphs show the performance of RRSE in validation and testing for all models and baselines. Validation results are shown in Figure \ref{subfig:SIR_val_fixed_l}, where baseline models (dotted lines) outperform MNNs (solid lines) for all durations. The MNNs are grouped in terms of their performance with a large gap to the baseline models. Moreover, the RF baseline model has the lowest RRSE across all durations, whereas PDC surprisingly has the highest RRSE of the MNNs. These validation results correspond well with the validation results from the previous experiment, where the baseline models outperformed the MNNs in modeling synthetic functions. The results of the out-of-distribution tests are shown in Figure \ref{subfig:SIR_test_fixed_l}, where the takeaway from validation to testing is quite similar to the previous experiment. The MNNs are also grouped together here, but have lower RRSE scores than the baseline models. For durations up to 12, the LR baseline model has the lowest RRSE narrowly beating the PDC. However, for the remaining durations, the RRSE of LR steadily increases and PDC has the lowest RRSE. PDC is closely followed by PDCLow and CCP, while PANN seems to consistently have a higher RRSE than the other MNNs. The RRSE of the baseline models exceeds one when a duration of 30 and greater is used. In contrast, this is only true for MNNs with a duration of 120. In general, it is observed that the performance of MNNs in validation is true to their performance in the out-of-distribution test. This is not the case for the baseline models, which have better performance in validation, but end up with poor performance in test for higher durations. 

\begin{figure}[ht]
    \centering
        \hspace{0.05\textwidth}
    \begin{subfigure}[t]{0.4\textwidth}
    \centering
    \includegraphics[width=\textwidth]{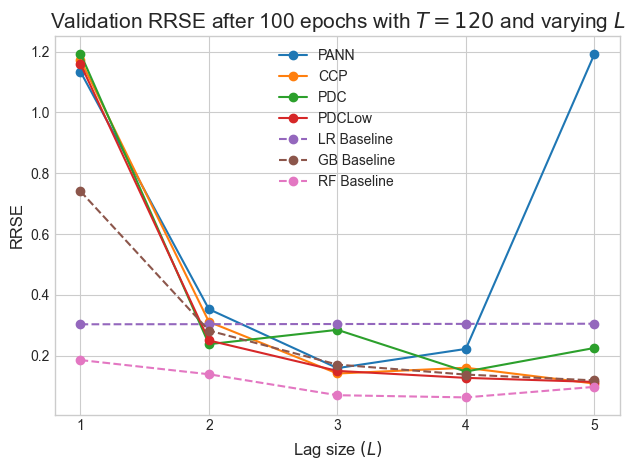}
    \caption{Validation performance where the MNNs are comparable to baseline models in RRSE.}
    \label{subfig:SIR_val_fixed_t}
    \end{subfigure}
    \hfill
    \begin{subfigure}[t]{0.4\textwidth}
    \centering
    \includegraphics[width=\textwidth]{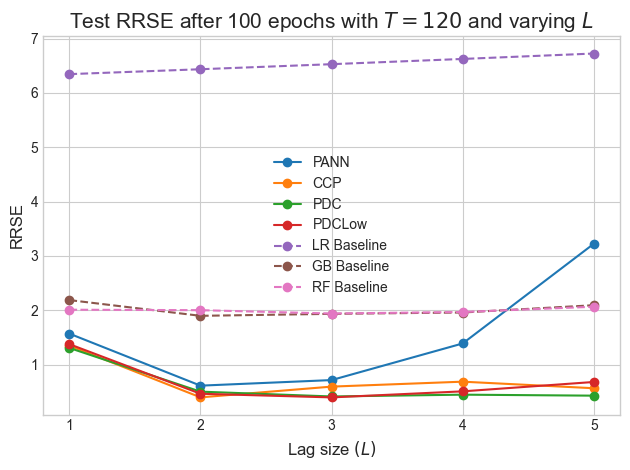}
    \caption{Testing performance where MNNs consistently outperform baseline models.}
    \label{subfig:SIR_test_fixed_t}
    \end{subfigure}
    \hspace{0.05\textwidth}
    \caption{Experiment with fixed duration and varying lag size. Baseline models are shown with dotted lines and MNNs are shown with solid lines. The y-axis shows the RRSE score, which means lower scores are better.}
    \label{fig:SIR_lag_results}
\end{figure}

In the second part of these results, the lag size is increased, while the duration is kept fixed at $T=120$. As the lag size increases, the polynomial order of the transition equations increases and the number of modeled time steps decreases, which should entail lower RRSE values, as there are fewer time steps for errors to accumulate in validation and testing. The results of increasing the lag size are visualized in Figure \ref{fig:SIR_lag_results}, where the validation results in Figure \ref{subfig:SIR_val_fixed_t} show a general pattern of decreasing RRSE scores as the lag size increases. The two exceptions are the LR baseline, which has nearly identical performance across all lag sizes, and PANN, which has an increasing RRSE using a lag size of four and five. The lowest RRSE score are seen for the baseline RF in all lag sizes, closely followed by the CCP, PDCLow, and the baseline GB with a lag size of five. The test results in Figure \ref{subfig:SIR_test_fixed_t} show how the MNNs outperform the baseline models in the out-of-distribution test in terms of RRSE, as in the first results of this experiment. PANN follows the same pattern as in validation, where the RRSE increases using a lag size of four and five. However, the RRSE score decreases for the remaining MNNs as the lag size increases. In contrast, the test RRSE of the baseline models is constant or increases slightly as the lag size increases. From these results, it is obvious that CCP, PDCLow, and PDC are capable of learning polynomials up to the 63rd order and generalize better than baseline models for these higher-order polynomials. 

\begin{figure}[ht]
\centering
    \hspace{0.05\textwidth}
    \begin{subfigure}[t]{0.4\textwidth}
    \centering
    \includegraphics[width=\textwidth]{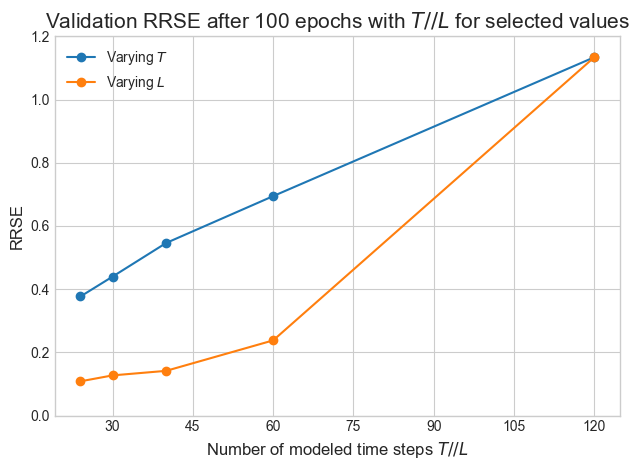}
    \caption{RRSE scores in validation showing that modeling higher-order polynomials (orange line) results in lower RRSE than modeling third-order polynomials (blue line).}
    \label{subfig:SIR_val_t_l}
    \end{subfigure}
    \hfill
    \begin{subfigure}[t]{0.4\textwidth}
    \centering
    \includegraphics[width=\textwidth]{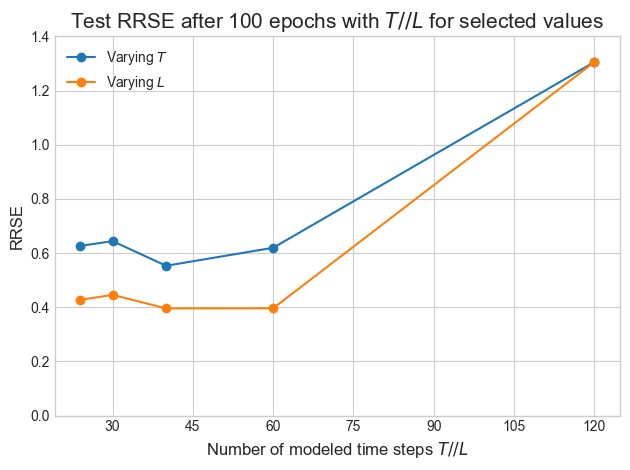}
    \caption{RRSE scores in testing. The difference between modeling higher-order polynomials (orange line) and third-order polynomials is less significant than in validation.}
    \label{subfig:SIR_test_t_l}
    \end{subfigure}
    \hspace{0.05\textwidth}
\caption{RRSE performance of MNNs as a function of the number of modeled time steps. The blue lines are a result of keeping lag size fixed while varying duration and vice versa for the orange line. Each point is the lowest RRSE score of the MNNs at a given value of $T//L$.}
\label{fig:SIR_sim_lag_results}
\end{figure}

% Comparison of simulations where the same number of time steps are modeled but polynomial order is higher for one
The final results of this experiment in Figure \ref{fig:SIR_sim_lag_results} combine the results of Figures \ref{fig:SIR_sequence_results} and \ref{fig:SIR_lag_results} to allow further interpretation of MNN performance. The two graphs show the lowest RRSE score of the MNNs at a specific number of modeled time steps. The blue line consists of the RRSE scores obtained using a fixed lag size $L=1$ and varying the duration by $T=[24, 30, 40, 60, 120]$, whereas the orange line comes from using a fixed duration $T=120$ and a variable lag size of $L=[5,4,3,2,1]$, that is, $T//L=[24, 30, 40, 60, 120]$. Therefore, RRSE is recorded at the same number of modeled time steps for both approaches, but the difference between the lines is the order of the polynomials that are modeled. For example, at 24 modeled time steps, a comparison can be made between modeling a third-order polynomial (blue) and a 63rd-order polynomial (orange). The RRSE scores at 120 modeled time steps come from the same run, but is shown for both lines for interpretability. In Figure \ref{subfig:SIR_val_t_l}, the validation results show that modeling higher-order polynomials is more desirable, as it results in lower RRSE scores than modeling third-order polynomials. The difference is most notable at 60 modeled time steps at a polynomial order of seven and least notable at 24 modeled time steps at a polynomial order of 63. For the testing, shown in Figure \ref{subfig:SIR_test_t_l}, the interpretation is similar to that in the validation but with a smaller difference when higher-order polynomials are modeled. However, the largest difference is still found at 60 modeled time steps. 

% should there be a wrap-up on results of this experiment?

\section{Discussion}\label{sec:discussion}
The results presented in this paper have a number of limitations, which stems from a lack of computational resources. First, the results come from a single experimental run, which ideally should be an aggregation of multiple runs with different seeds. Therefore, the robustness of the results and conclusions is unknown. Second, hyperparameters are selected manually and used universally for all models and runs, where automated hyperparameter tuning for each model and learning problem is more desirable. A third limitation related to using universal hyperparameters is that models are trained for the same number of epochs despite varying greatly in number of parameters. Ideally, each model should be trained until the validation loss has converged. However, better performance in validation does not necessarily lead to better performance in out-of-distribution tests. 

Out-of-distribution is a central concept in this work, as it is used to test generalization capabilities of the models. In this aspect, nonoverlapping training and test distributions have been considered sufficient, but it has not been quantified how far these distributions should be apart from each other for testing to be considered out-of-distribution. This is especially a limitation to the experiment in simulation metamodeling, as distributions for multiple parameters have to be considered. Furthermore, for this experiment, polynomials of very high order are modeled, leading to significantly different training times of the metamodels. Although these training times could influence whether MNNs are more suitable metamodels than the baseline models considered, they have not been reported in this work. A final limitation is the absence of kriging or Gaussian processes as a baseline model, given its importance and popularity in the metamodeling literature.

% Future work
For future work, it would be interesting to experiment with additional variations of MNN architectures. An obvious extension to the MNN models considered in this work, would be a deep version of CCP implemented by adding hidden layers, in the same spirit as the $V$ layers of PDC. This would enable comparisons of whether deeper models perform better in generalizing polynomials and simulation metamodeling. Moreover, the proposed simulation metamodeling approach should be applied for more simulation models, and preferably for simulations that are computationally expensive, as opposed to the SIR simulation considered in this work.

\section{Conclusion}\label{sec:conclusion}
% Highlights the importance of polynomials in simulations and their out-of-distribution generalization
% MNNs learning polynomials in-distribution and generalizing out-of-distribution
    % Can struggle if output has a large range
% Showing how increasing the size of time steps increases the order of polynomial describing time step updates
% Simulation metamodeling approach
    % Works with any machine learning model in simulation where time step updates are described by polynomials
% Metamodeling approach further shows the inductive bias of MNNs preferring learning higher-order polynomials as opposed to lower-order polynomials
    % Also beats baseline models
This thesis highlights the importance of being able to learn and generalize polynomials for simulation metamodels. For this purpose, MNN architectures are used, which can raise input to higher-order output through multiplication. PANN, CCP, and PDC were collected from the literature and repurposed, while PDCLow was proposed based on these. 
Experiments on polynomial learning showed that MNNs are capable of generalizing out-of-distribution for sixth-order polynomials, while struggling for higher-order polynomials. From this experiment alone, it could not be determined whether this was the case due to the high order of the polynomial or the wide output range incurred from increasing the polynomial order, which increased the chances of numerical overflow in out-of-distribution tests. 
To account for this, the second experiment modeled synthetic functions with constrained input domains, where it was evident that the width of the output range, rather than the polynomial order, influenced the ability of MNNs to learn and generalize a polynomial. Additionally, the validation and out-of-distribution testing performance of MNNs was more similar than that for the baseline models. This was further confirmed in the experiment that metamodeled an epidemiology simulation. 

The proposed simulation metamodeling approach showed how increasing the size of the time steps entailed approximating higher-order polynomials for simulations with time step updates described by polynomials. A useful property of this approach is that the number of time steps, where approximation errors can accumulate, is lowered while simulating the same time interval. The third experiment exemplified this by showing decreasing RRSE scores for MNNs as a result of increasing the lag size. Furthermore, the third experiment showed better performance for MNNs approximating higher-order polynomials as opposed to lower-order polynomials, which supports the presence of an inductive bias in MNNs towards learning and generalizing polynomials. An inductive bias for generalizing higher-order polynomials is critical to simulation metamodels, as it allows simulations further ahead in time in the same amount of time steps.

\bibliography{bibliography}

\newpage
\appendix
\section{Appendix}\label{sec:appendix}
\subsection{Preliminary Experiments with Arbitrary Polynomials}\label{sec:appendix-prelim-exp}
\begin{figure}[ht]
    \centering
    \begin{subfigure}[t]{0.2\textwidth}
    \centering
    \includegraphics[width=\textwidth]{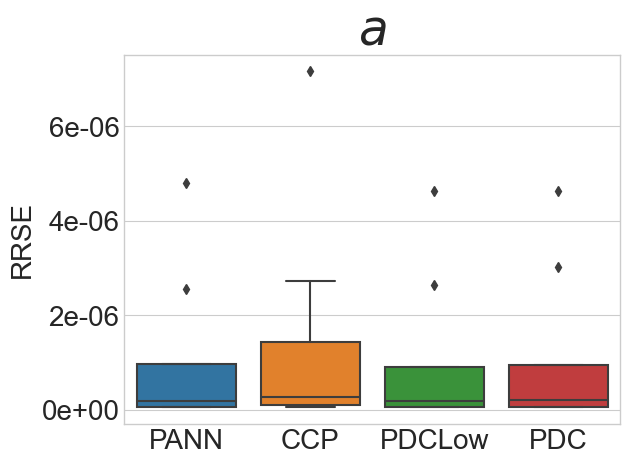}
    \end{subfigure}
    \hfill
    \begin{subfigure}[t]{0.2\textwidth}
    \centering
    \includegraphics[width=\textwidth]{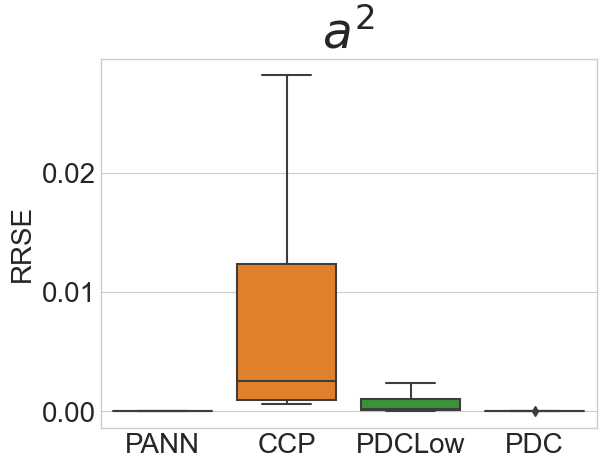}
    \end{subfigure}
    \hfill
    \begin{subfigure}[t]{0.2\textwidth}
    \centering
    \includegraphics[width=\textwidth]{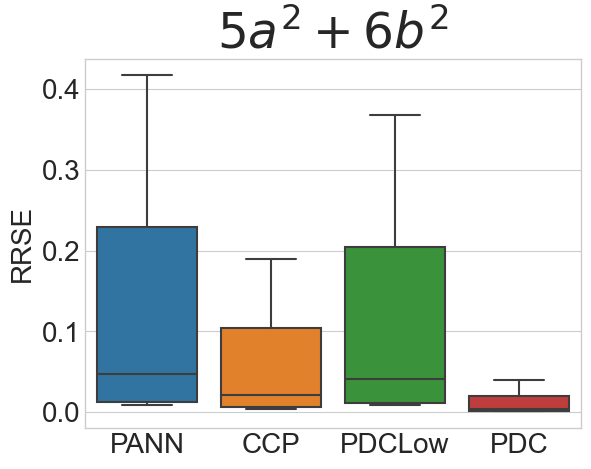}
    \end{subfigure}
    \hfill
    \begin{subfigure}[t]{0.2\textwidth}
    \centering
    \includegraphics[width=\textwidth]{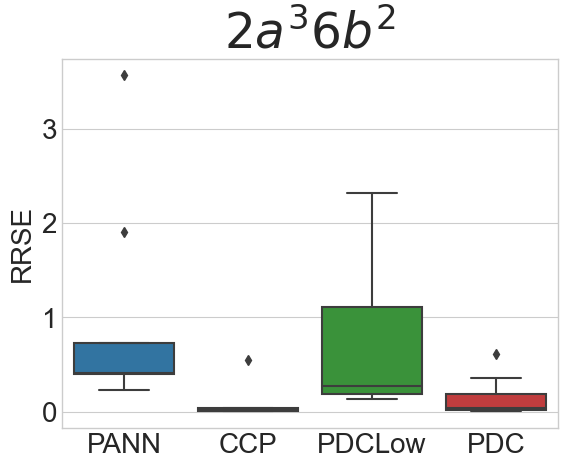}
    \end{subfigure}
    \hfill
    \begin{subfigure}[t]{0.2\textwidth}
    \centering
    \includegraphics[width=\textwidth]{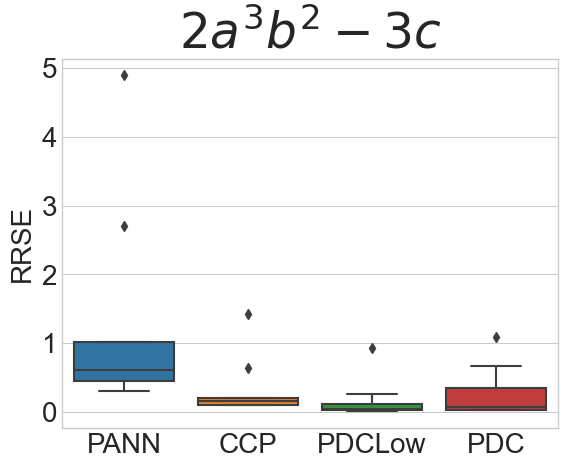}
    \end{subfigure}
    \hfill
    \begin{subfigure}[t]{0.2\textwidth}
    \centering
    \includegraphics[width=\textwidth]{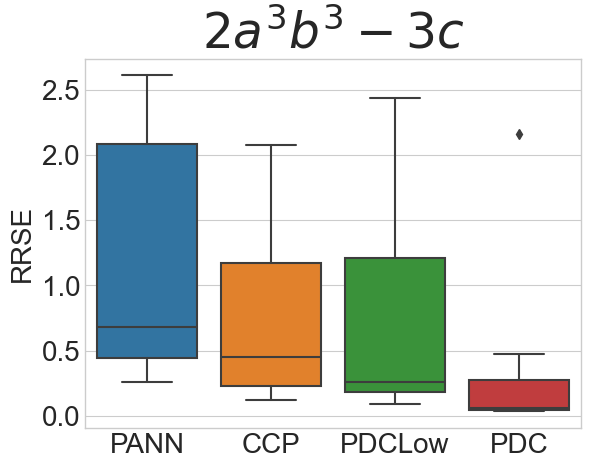}
    \end{subfigure}
    \hfill
    \begin{subfigure}[t]{0.2\textwidth}
    \centering
    \includegraphics[width=\textwidth]{Pictures/Results/2a2_b2_c2-3d.png}
    \caption{Shown in paper.}
    \label{appendix:2a2_b2_c2-3d}
    \end{subfigure}
    \hfill
    \begin{subfigure}[t]{0.2\textwidth}
    \centering
    \includegraphics[width=\textwidth]{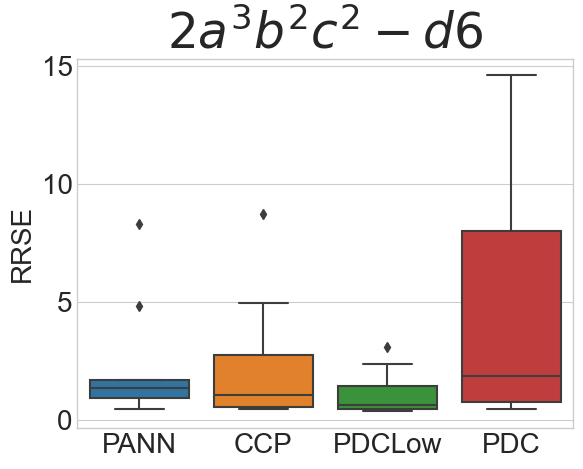}
    \end{subfigure}
    \hfill
    \begin{subfigure}[t]{0.2\textwidth}
    \centering
    \includegraphics[width=\textwidth]{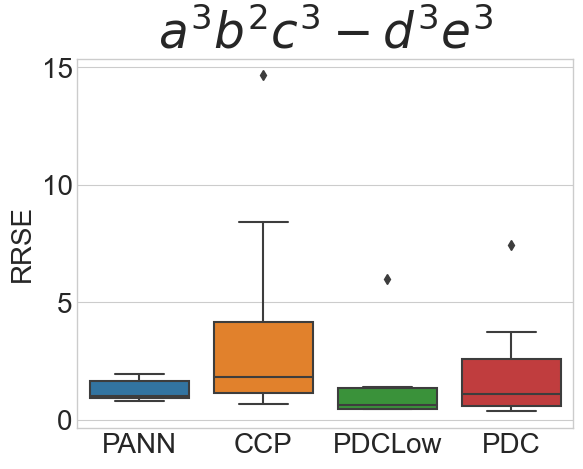}
    \end{subfigure}
    \hfill
    \begin{subfigure}[t]{0.2\textwidth}
    \centering
    \includegraphics[width=\textwidth]{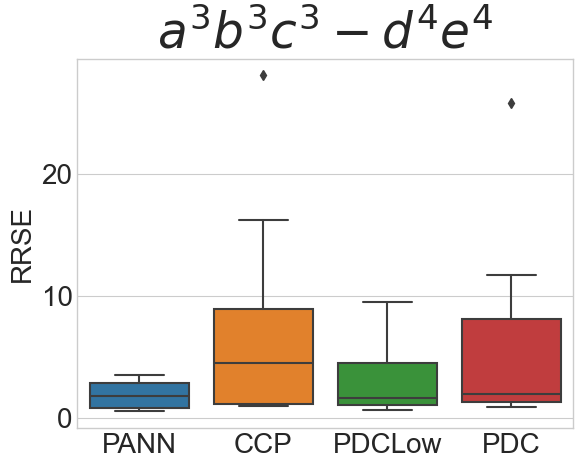}
    \end{subfigure}
    \hfill
    \begin{subfigure}[t]{0.2\textwidth}
    \centering
    \includegraphics[width=\textwidth]{Pictures/Results/a3_b2_c2_d3-e5_f5.png}
    \caption{Shown in paper.}
    \label{appendix:a3_b2_c2_d3-e5_f5}
    \end{subfigure}
    \hfill
    \begin{subfigure}[t]{0.2\textwidth}
    \centering
    \includegraphics[width=\textwidth]{Pictures/Results/a_b_c_d_e_f_g.png}
    \caption{Shown in paper.}
    \label{appendix:a_b_c_d_e_f_g}
    \end{subfigure}
    \hfill
    \begin{subfigure}[t]{0.2\textwidth}
    \centering
    \includegraphics[width=\textwidth]{Pictures/Results/a_b_c+d-e-f-g.png}
    \caption{Shown in paper.}
    \label{appendix:a_b_c+d-e-f-g}
    \end{subfigure}
    %\hfill
    \begin{subfigure}[t]{0.2\textwidth}
    \centering
    \includegraphics[width=\textwidth]{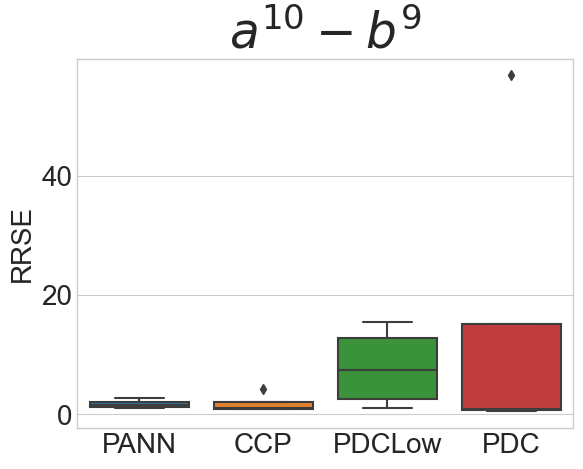}
    \end{subfigure}
    \hfill
    \caption{Test results from learning arbitrary polynomials with MNNs. Models are tested on nine distributions where eight are out-of-distribution from the training.}
    \label{fig:appendix-arbitrary-full}
\end{figure}

Preliminary experiments with arbitrary polynomials are conducted to test the viability of learning polynomials with MNNs in-distribution and out-of-distribution. Selected results are shown in the paper while the rest are shown here. The validation results from the use of a Gaussian $\mathcal{N}(0,5)$ training distribution are shown in Table \ref{tab:appendix-prelim-exp}. The test results are visualized as box plots from the RRSE scores of nine test distributions, which are Gaussian distributions with mean and standard deviation composed of all combinations between $\mu=[-50,0,90]$ and $\sigma=[1, 5, 25]$. The results for all polynomials are shown in Figure \ref{fig:appendix-arbitrary-full}.
\begin{table}[h]
  \centering
  \caption{Validation RRSE for all arbitrary polynomials and models using samples from a Gaussian $\mathcal{N}(0,5)$ distribution. Bold indicates the lowest RRSE score for a polynomial.}
  \label{tab:appendix-prelim-exp}
  \begin{tabular}{|l|l|l|l|l|}
    \hline
    \textbf{Polynomial} & \textbf{PANN} & \textbf{CCP} & \textbf{PDCLow} & \textbf{PDC} \\ \hline
    $a$ & 5.555e-08 & 5.216e-08 & 5.471e-08 & \textbf{5.094e-08} \\[0.5ex] \hline
    $a^2$ & \textbf{9.245e-08} & 9.067e-04 & 7.731e-05 & 9.453e-08 \\[0.5ex] \hline
    $5a^2 + 6b^2$ & 0.008 & 0.005 & 0.008 & \textbf{0.001} \\[0.5ex] \hline
    $2a^3 6b^2$ & 0.400 & 0.045 & 0.220 & \textbf{0.014} \\[0.5ex] \hline
    $2a^3 b^2 - 3c$ & 0.457 & 0.193 & 0.038 & \textbf{0.027} \\[0.5ex] \hline
    $2a^3 b^3 - 3c$ & 0.722 & 0.487 & 0.206 & \textbf{0.045} \\[0.5ex] \hline
    $2a^2 b^2 c^2 - 3d$ & 0.834 & \textbf{0.349} & 0.350 & 0.387 \\[0.5ex] \hline
    $2a^3 b^2 c^2 - d^6$ & 0.861 & 0.522 & \textbf{0.451} & 0.465 \\[0.5ex] \hline
    $a^3 b^2 c^3 - d^3 e^3$ & 1.741 & 1.229 & 1.368 & \textbf{1.167} \\[0.5ex] \hline
    $a^3 b^3 c^3 - d^4 e^4$ & 1.904 & \textbf{0.932} & 1.295 & 1.090 \\[0.5ex] \hline
    $a^3 b^2 c^2 d^3 - e^5 f^5$ & 4.690 & 2.672 & \textbf{1.862} & 2.249 \\[0.5ex] \hline
    $a b c d e f g$ & 2.561 & 1.636 & \textbf{1.510} & 1.522 \\[0.5ex] \hline
    $a b c + d - e - f - g$ & 0.046 & 0.042 & \textbf{0.003} & 0.059 \\[0.5ex] \hline
    $a^{10} - b^9$ & 1.237 & 0.975 & 1.050 & \textbf{0.793} \\[0.5ex] \hline
  \end{tabular}
\end{table}

\subsection{Synthetic Functions}\label{sec:appendix-opt-func}
\begin{figure}[ht]
    \centering
    \begin{subfigure}[b]{0.3\textwidth}
    \centering
    \includegraphics[width=\textwidth]{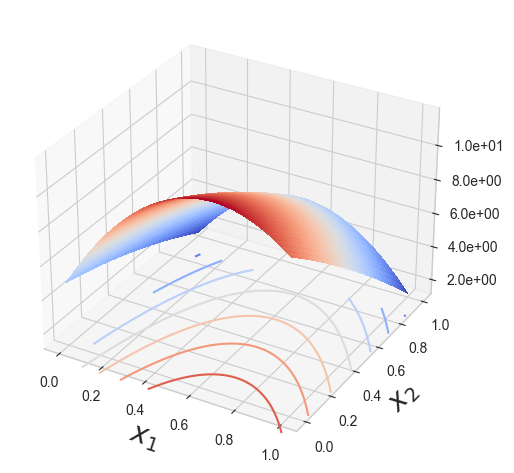}
    \caption{Currin function\cite{Currin-function}. No reported optimization success rate.}
    \label{subfig:Currin}
    \end{subfigure}
    \hfill
    \begin{subfigure}[b]{0.3\textwidth}
    \centering
    \includegraphics[width=\textwidth]{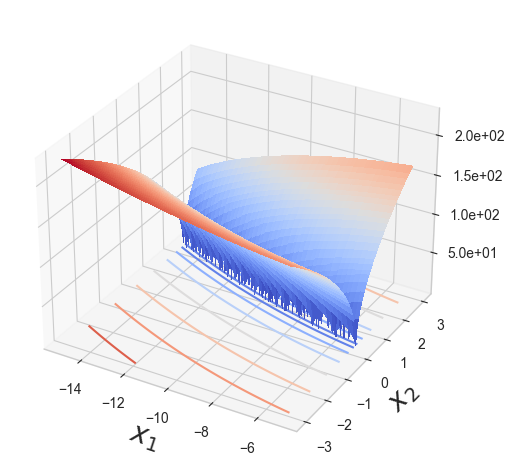}
    \caption{Bukin 6 function \cite{Bukin-Function}. Successful optimization rate of 0.31\% \cite{Gavana}}
    \label{subfig:Bukin6}
    \end{subfigure}
    \hfill
    \begin{subfigure}[b]{0.3\textwidth}
    \centering
    \includegraphics[width=\textwidth]{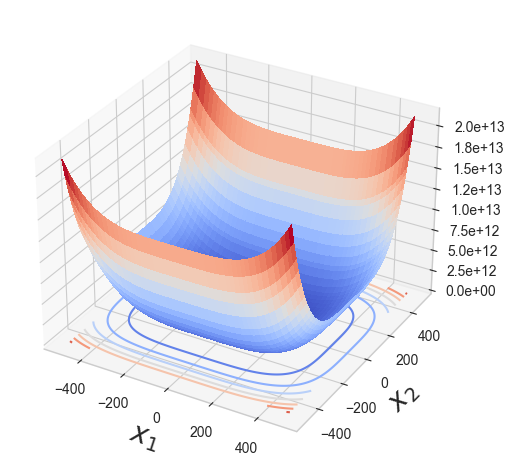}
    \caption{Price 3 function \cite{Price3-Function}. Successful optimization rate of 70.77\%}
    \label{subfig:Price3}
    \end{subfigure}
    \hfill
    \begin{subfigure}[b]{0.3\textwidth}
    \centering
    \includegraphics[width=\textwidth]{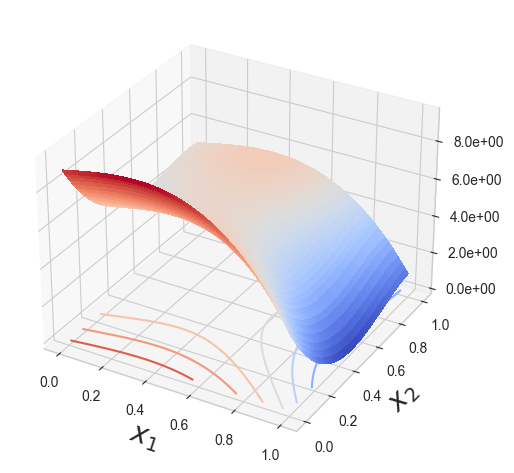}
    \caption{Lim polynomial \cite{Lim-Polynomial}}
    \label{subfig:Lim}
    \end{subfigure}
    \hfill
    \begin{subfigure}[b]{0.3\textwidth}
    \centering
    \includegraphics[width=\textwidth]{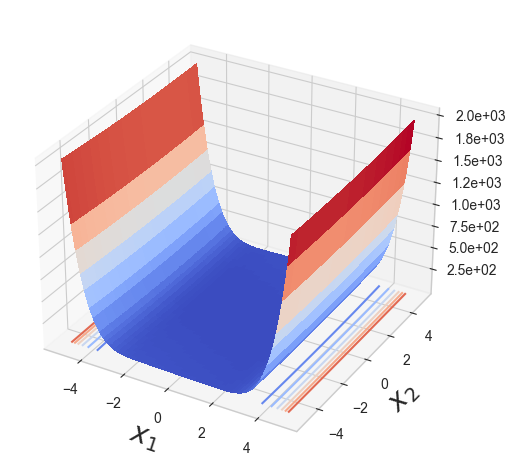}
    \caption{Camel Three Hump function \cite{Camel-Function}. Successful optimization rate of 95.54\%}
    \label{subfig:CamelThreeHump}
    \end{subfigure}
    \hfill
    \begin{subfigure}[b]{0.3\textwidth}
    \centering
    \includegraphics[width=\textwidth]{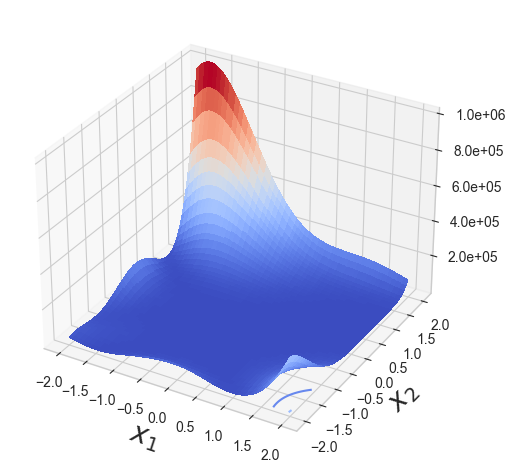}
    \caption{Goldstein-Price function \cite{GoldsteinPrice-Function}. Successful optimization rate of 96.15\%}
    \label{subfig:GoldsteinPrice}
    \end{subfigure}
    \begin{subfigure}[b]{0.3\textwidth}
    \centering
    \includegraphics[width=\textwidth]{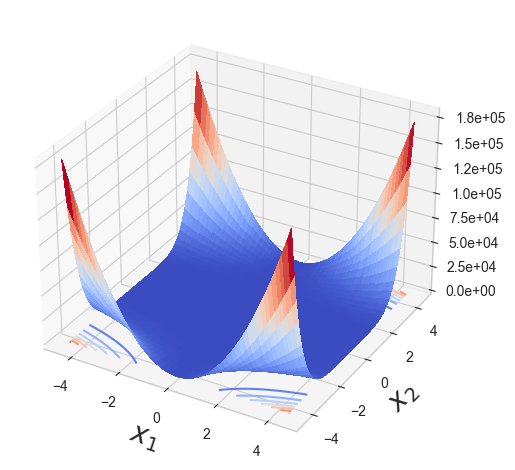}
    \caption{Beale function \cite{BenchmarkOptimizationProblems}. Successful optimization rate of 96.69\%}
    \label{subfig:Beale}
    \end{subfigure}
    \caption{Contour surface plots of synthetic functions with two variables. The values along the z-axis are the output of $f(x_1, x_2)$, where the function $f$ is the synthetic function as described in the caption of each subplot.}
    \label{fig:appendix-optimization-functions-viz}
\end{figure}

The synthetic functions of Table \ref{tab:opt-func} are shown in Equations \ref{eq:Currin}-\ref{eq:Goldstein-Price} and the ones with two input variables are visualized as surface plots in Figure \ref{fig:appendix-optimization-functions-viz}. The first synthetic function has two variables, a low polynomial order of two, and a narrow input domain for its variables. The Currin function is considered the simplest of the synthetic functions, although it does not have the difficulty reported \cite{Gavana}. The second synthetic function is reported to be the toughest, with an overall success of 0.31\% using a wide range of optimizers. Although the Bukin function uses both the square root and absolute value operators, it is included to determine whether functions that are deemed extremely difficult to optimize are also difficult to learn out-of-distribution. The function is visualized in Figure \ref{subfig:Bukin6}, where the function plane is similar to that of a paper plane. The third synthetic function, the Price 3 function, has a reported optimization success of 70.77\%, but the output of the function ranges greatly in magnitude, as seen in Figure \ref{subfig:Price3}. Additionally, the input domain is the widest of all synthetic functions. The Dette-Pepelyshev function is the fourth synthetic function, has a polynomial order of four, and a narrow input domain. Like the Bukin 6 function, this function uses the square root operator, but is much simpler. The fifth function, the Colville function, has the largest number of variables in the synthetic functions. It has a reported optimization success of 59.31\%, which is the second lowest of the synthetic functions used in this experiment. The sixth synthetic function, the Lim polynomial function, has a polynomial order of five, but is considered one of the simpler synthetic functions. The input domain is narrow, and the function hyperplane does not have unexpected behavior in the extremes of the input domain. The ThreeHumpCamel function is the seventh synthetic function and is shown in Figure \ref{subfig:CamelThreeHump}. It is not evident from the figure, but the three humps come from the three local minima that occur around (0,0). However, this cannot be viewed from the figure, as the fluctuations are much smaller than the overall magnitude of the function output. It is a relatively simple function with an optimization success of 95.54\%, although extreme behavior occurs in the extremes of the input domain. Likewise, the eighth synthetic function, the Beale function, has extreme behavior in the extremes of the input domain compared to the rest of the input domain. Despite this, Beale is reported to be one of the least tough functions with an optimization success of 96.69\%. The final and ninth synthetic function is visualized in Figure \ref{subfig:GoldsteinPrice}. The function shows extreme behavior in one corner of its hyperplane and is comprised of many terms compared to the other synthetic functions. It is still a simple function with an optimization success of 96.15\%.

\begin{align}
f_{Currin}(x) &= 4.9 + 21.15x_1 -2.17x_2 - 15.88x_1^2 - 1.38x_2^2 -5.26x_1 x_2 \label{eq:Currin}\\
f_{Bukin6}(x) &= 100\sqrt{|x_2-0.01x_1^2|} + 0.01|x_1+10| \label{eq:Bukin6}\\
f_{Price3}(x) &= 100(x_2-x_1^2)^2 + 6\left(6.4(x_2-0.5)^2-x_1-0.6\right) \label{eq:Price3}\\
f_{D\&P}(x) &= 4(x_1-2+8x_2-8x_2^2)^2 + (3-4x_2)^2 + 16 \sqrt{x_3+1}(2x_3-1)^2 \\
f_{Colville}(x) &= 100(x_1-x_2^2)^2 + (1-x_1)^2 + 90(x_4-x_3^2)^2 + (1-x_3^2) \\
& \quad + 10.1((x_2-1)^2 + (x_4-1)^2)+19.8(x_2-1)(x_4-1) \nonumber \\
f_{Lim}(x) &= 9 + \frac{5}{2}x_1 - \frac{35}{2}x_2 + \frac{5}{2}x_1 x_2 + 19x_2^2 - \frac{15}{2}x_1^3 - \frac{5}{2}x_1 x_2^2 - \frac{11}{2}x_2^4 + x_1^3 x_2^2 \\
f_{Camel}(x) &= 2x_1^2 − 1.05x_1^4 + \frac{x_1^6}{6} + x_1 x_2 + x_2^2 \label{eq:Camel}\\
f_{Beale}(x) &= (1.5-x_1+x_1 x_2)^2 + (2.25 - x_1 + x_1 x_2^2)^2 + (2.625 - x_1 + x_1 x_2^3)^2 \\
f_{Goldstein-Price} &= \left[1+(x_1+x_2+1)^2(19-14x_1 + 3x_1^2 - 14x_2 + 6x_1 x_2 + 3x_2^2)\right] \label{eq:Goldstein-Price}\\
& \quad \,\left[30+(2x_1-3x_2)^2(18-32x_1 + 12x_1^2 + 48x_2-36x_1x_2 + 27x_2^2)\right] \nonumber
\end{align}

\iffalse
\begin{figure}[ht]
    \centering
    \includegraphics[width=0.5\textwidth]{Pictures/Appendix/Currin.png}
    \caption{Currin function \cite{Currin-function}.}
    \label{fig:Currin}
\end{figure}

\begin{figure}[ht]
    \centering
    \includegraphics[width=0.5\textwidth]{Pictures/Appendix/Bukin06.png}
    \caption{Bukin 6 function \cite{Bukin-Function}.}
    \label{fig:Bukin}
\end{figure}

\begin{figure}[ht]
    \centering
    \includegraphics[width=0.5\textwidth]{Pictures/Appendix/Bukin06.png}
    \caption{Bukin 6 function \cite{Bukin-Function}.}
    \label{fig:Bukin}
\end{figure}
\fi

\subsection{Experiment Metrics}\label{sec:appendix-metrics}
Metrics are recorded for all experiments in the Python TorchMetrics library \citet{TorchMetrics} and root relative squared error (RRSE) as defined in the GeneXproTools 4.0 software \citet{RRSE}. Training and validation loss is implemented as the mean squared error (MSE) of the error between true values and predictions, as shown in Equation \ref{eq:mse}. For diagnostics during model training, the $R^2$ score and RRSE are logged in validation after each epoch. The $R^2$ score represents the proportion of variance of the target $y$ explained by the independent variable(s) in the model. Therefore, $R^2$ cannot exceed one, as the independent variable(s) cannot explain more variance than there is. This is consistent with the fact that a perfect model would result in the numerator in Equation \ref{eq:r2} being zero and thereby subtracting zero from one. $R^2$ should not be below zero, but it is possible if the model is worse than predicting the average of the target $y$. RRSE is largely similar to $R^2$ as seen by comparing Equations \ref{eq:r2} and \ref{eq:appendix-rrse}. A subtle but big difference is that the target $y$ is subtracted from the prediction $\hat{y}$ in the RRSE and not the other way around for $R^2$. RRSE ranges between zero and infinity, where being closer to zero indicates better performance in contrast to $R^2$. In addition to $R^2$ and RRSE, models are evaluated in the out-of-distribution tests with mean absolute error (MAE) as seen in Equation \ref{eq:mae}.
\begin{align}
    \text{MSE} &= \frac{1}{n}\sum_{i=1}^n (y_i-\hat{y_i})^2 \label{eq:mse}\\
    R^2 &= 1-\frac{\sum_{i=1}^n(y_i - \hat{y_i})^2}{\sum_{i=1}^n(y_i-\Bar{y})^2}\label{eq:r2} \\
    \text{RRSE} &= \sqrt{\frac{\sum_{i=1}^n (\hat{y}_i-y_i)^2}{\sum_{i=1}^n (y_i- \Bar{y})^2}} \label{eq:appendix-rrse} \\
    \text{MAE} &= \frac{1}{n} \sum_{i=1}^n |y_i-\hat{y_i}| \label{eq:mae}
\end{align}

Absolute metrics, such as MSE and MAE, allow meticulous comparisons of model performance within the same polynomial and data distribution. MSE is recorded to be able to directly interpret the learning process of the neural networks. However, the MSE is on a different scale than the data, and it can be hard to distinguish unsatisfactory MSE values from satisfactory ones. This is the main motivation behind including MAE since this is directly comparable to the scale of the data. Unfortunately, comparisons with absolute metrics cannot be made across polynomials and data distributions, which introduced the need for relative metrics in this project. As a result, the relative metrics, $R^2$ and RRSE, are included for this reason. For really simple learning problems, it can be relatively easy to achieve a perfect $R^2$ of one. If multiple models reach this score, it is impossible to separate them in evaluation. Fortunately, it is rare to train a model that has perfect predictions, which means that RRSE might be more appropriate when comparing great-performing models. Absolute and relative metrics complement each other and are used on a shifting basis. 

%\subsection{Learning Objectives}
%This paper has been written as part of a thesis at the Technical University of Denmark. It is common practice to have learning objectives that highlight the objectives that the author is expected to achieve in a successful project. The learning objectives are formulated as:
%\begin{itemize}
%    \item Define polynomials and their usefulness in modeling real-world phenomena and processes through simulation
%    \item Describe the relationship between neural networks and polynomials
%    \item Conduct small-scale experiments with neural networks that learn polynomials
%    \item Formulate neural network architectures with inductive bias for learning polynomials
 %   \item Demonstrate performance of formulated architectures compared to baseline machine learning models
 %   \item Apply formulated architectures as a simulation metamodel
 %   \item Organize code in an efficient way for reproducibility and shareability
 %   \item Be able to present findings and methodology clearly in a report
%\end{itemize}

\end{document}